\documentclass[runningheads]{llncs}

\usepackage{accv}

\usepackage{accvabbrv}

\usepackage{graphicx}
\usepackage{booktabs}
\usepackage{multirow}
\usepackage{wrapfig}
\usepackage{tabularx}
\usepackage{tcolorbox}
\usepackage[table]{xcolor}
\usepackage{rotating}
\usepackage{dsfont}

\usepackage[accsupp]{axessibility}  

\newif\ifsupp
\supptrue

\usepackage{hyperref}

\usepackage{orcidlink}

\begin{document}

\title{A Canary in the Neural Mine: Catching an Image Classifier's Silent, Confident Failures} 

\titlerunning{Catching an Image Classifier's Silent, Confident Failures}

\author{Wonseok Shin\orcidlink{0009-0007-3582-763X} \and
Songkuk Kim\thanks{Corresponding author.}\orcidlink{0000-0003-4147-4627}}

\authorrunning{W.~Shin et al.}

\institute{Yonsei University, Seoul 03722, Republic of Korea \\
\email{\{wonseok.shin,songkuk\}@yonsei.ac.kr}}

\maketitle

\begin{abstract}

Deployed vision models auto-accept confident predictions, so high-confidence errors propagate silently. Maximum Softmax Probability (MSP), the standard failure predictor, has a structural blind spot: as a scalar read from the classifier head, it is first-order blind to feature variation orthogonal to its confidence gradient. Through a factorial analysis of feature-space detectors, we identify a conditional-redundancy principle explaining why they fail to complement MSP: the class-separability geometry they measure is already encoded in MSP and collapses under fusion. The signal that survives is class-conditional typicality on the off-manifold residual---and because this residual's distribution varies by backbone, from near-Gaussian to heavy-tailed, it must be estimated non-parametrically. We instantiate these findings as CANARY, a training-free detector that computes a per-class kNN distance on the PCA residual of the penultimate feature, percentile-calibrates it on validation data, and fuses it with MSP through a single validation-tuned weight. CANARY's benefit follows a headroom law: gains grow as baseline MSP quality falls, so a single clean-validation number forecasts the improvement. With every parameter selected on in-distribution data alone, CANARY matches or improves 382 of 385 shift cells (e-AURC; 379 on AUROC) across five backbones, with worst-case regression within seed noise ($-0.0002$ AUROC) and the largest gains on natural distribution shifts (ImageNet-A/R, ObjectNet)---where deployed models are most exposed.


\end{abstract}

\section{Introduction}
\label{sec:intro}

Deployed vision models operate under selective classification: they auto-accept predictions when confident and abstain when uncertain. The safety of this pipeline relies entirely on accurate failure prediction, defined as the model's ability to rank correct predictions above incorrect ones. When ranking fails at high confidence, errors propagate silently into deployment.

Maximum Softmax Probability (MSP) remains the canonical baseline for failure prediction, but it suffers from a structural blind spot. By collapsing the $D$-dimensional feature space to a scalar confidence, MSP is first-order insensitive to an entire $(D-1)$-dimensional hyperplane of feature variation---on ViT-B/16, 767 of 768 dimensions (Section~\ref{sec:analysis-blindspot}). Feature-space methods such as ViM and the Mahalanobis family exploit exactly this discarded geometry---yet, designed for OOD detection, they consistently underperform MSP when transplanted to failure prediction (Section~\ref{sec:related-work}).

\begin{figure}[tb]
  \centering
  \includegraphics[width=\textwidth]{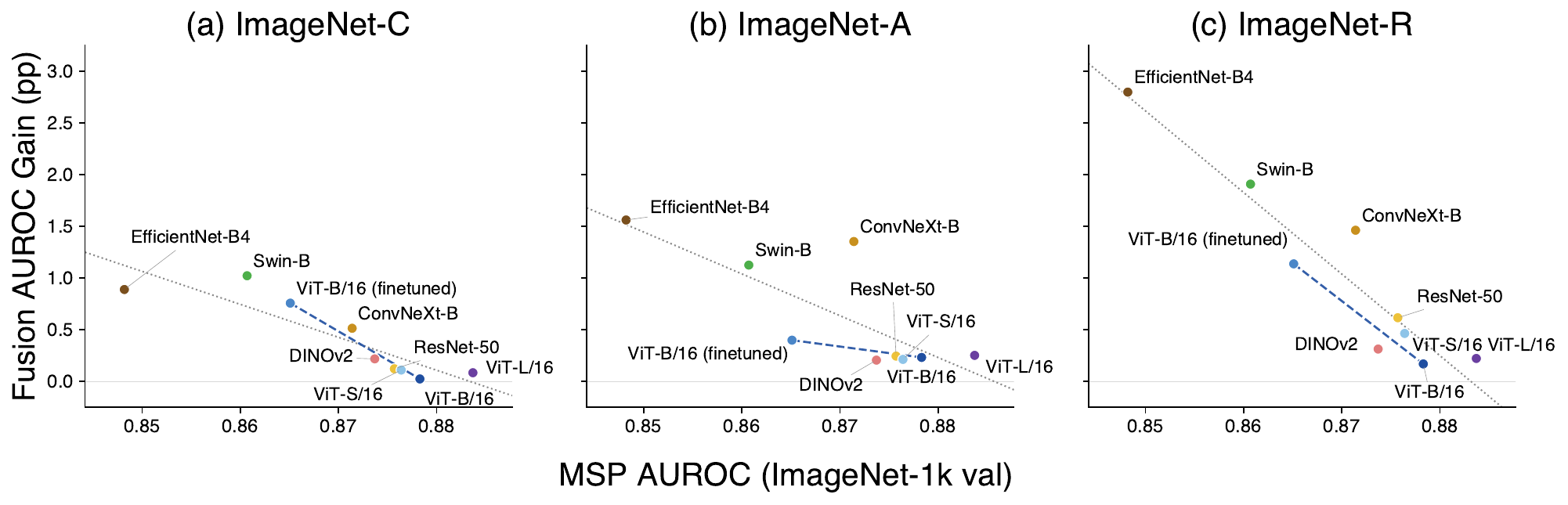}
  \caption{\textbf{The Headroom Law across shift families.} CANARY's fusion AUROC gain (y-axis) is inversely related to the backbone's baseline MSP AUROC on the clean ImageNet-1k validation set (x-axis). This predictable negative-slope relationship holds consistently across both synthetic ((a) ImageNet-C) and natural shifts ((b) ImageNet-A, (c) ImageNet-R), with natural shifts exhibiting the most substantial gains. Dashed line: a same-architecture control (two ViT-B/16 fine-tunes) isolating MSP quality as the driver of the gain (Section~\ref{sec:experiments}).}
  \label{fig:headroom-law}
\end{figure}

To understand exactly what information in the feature space complements MSP, we decompose the design space of feature-based failure prediction into a factorial experiment over  three axes: projection basis, distance metric, and reference conditioning.  This analysis reveals a conditional-redundancy principle: the class-separability geometry that most feature detectors measure is redundant with MSP under fusion---the classifier head has already extracted those directions. What survives fusion is off-manifold typicality, and it imposes two requirements: the score must be non-parametric, because the residual's distribution varies by backbone, and class-conditional, because global typicality breaks under natural shifts (Section~\ref{sec:analysis}).

We introduce CANARY, a minimal, training-free instantiation of these findings: a per-class kNN distance on the PCA residual of the penultimate feature, percentile-calibrated and fused with MSP through a single validation-tuned weight. Because CANARY isolates precisely the signal MSP lacks, its benefit is predictable: fusion gains grow as baseline MSP quality falls (Figure~\ref{fig:headroom-law}), so a single clean-validation number forecasts the gain before deployment---small where MSP is already a near-ceiling ranker, largest on weak backbones and natural shifts. A same-architecture control isolates the driver: two ViT-B/16 fine-tunes differing only in MSP quality show gains of $+0.42$ vs.\ $+0.08$ pp, ruling out architectural confounds.

This predictability yields a concrete operational property---empirically, CANARY is never worse than the MSP it augments. Across 385 shift cells on five backbones, it matches or improves 382 cells (e-AURC; 379 AUROC); the exceptions (3 on e-AURC, 6 on AUROC) lie within seed noise (worst case $-0.0002$ AUROC). Its gains concentrate on natural, real-world shifts (ImageNet-A, ImageNet-R, ObjectNet) rather than synthetic corruptions.

\vspace{0.7em}
\noindent \textbf{Contributions:}
\begin{itemize}
\item \textbf{Conditional-Redundancy Principle.} A factorial analysis over basis, distance, and conditioning showing that class-separability geometry is redundant with MSP under fusion; the complementary signal is non-parametric, class-conditional, off-manifold typicality (Section~\ref{sec:analysis}).

\vspace{0.5em}
\item \textbf{Headroom law.} Fusion gains decrease predictably with baseline MSP quality, isolated from architecture by a same-architecture control and validated across nine ImageNet backbones (including self-supervised DINOv2), a different label space (CIFAR-100), and a held-out covariate shift (ObjectNet) (Section~\ref{sec:experiments}).

\vspace{0.5em}
\item \textbf{CANARY.} A training-free, post-hoc detector realizing the blueprint; it matches or exceeds MSP on 382 of 385 shift cells with worst-case regression within seed noise, with the largest gains on natural shifts  (Sections~\ref{sec:method}--\ref{sec:experiments}).

\end{itemize}

\section{Related Work}
\label{sec:related-work}

\noindent \textbf{Logit- and Activation-Based Detection.} MSP~\cite{hendrycks2017baseline} remains the canonical baseline for failure prediction. Logit-space corrections, such as MaxLogit~\cite{Hendrycks2022ScalingOD}, Energy~\cite{liu2020energy}, Doctor~\cite{Granese21}, and GEN~\cite{GEN-Liu23}, rely purely on the information already exposed by the linear classifier head, inherently inheriting MSP's structural blind spot. Because any score read from the logit vector alone is invariant to feature variation orthogonal to the classifier head, it cannot retroactively distinguish confident errors from correct predictions; extensive evaluations confirm none consistently outperform MSP~\cite{Jaeger23}. Activation-shaping heuristics (\eg, ReAct~\cite{Sun21}, ASH~\cite{Djurisic23}, DICE~\cite{Sun22b}) modify features before logit computation. However, being calibrated primarily on CNN statistics, they suffer from severe architecture sensitivity. When transplanted to structurally different models like Vision Transformers, their performance often degrades sharply.

\vspace{0.7em}
\noindent \textbf{Feature-Space Detection.} To bypass the logit bottleneck, methods like Mahalanobis~\cite{Lee18}, Deep kNN~\cite{Sun22}, ViM~\cite{ViM-Wang22}, and Kernel PCA~\cite{NEURIPS2024_f2543511} leverage broader feature geometry. However, these methods were designed specifically for out-of-distribution (OOD) detection. Detecting a confidently misclassified in-distribution sample is structurally distinct from flagging an anomalous OOD input. Consequently, when transplanted to failure prediction, they consistently---and often substantially---underperform the MSP baseline. Furthermore, heuristic parameter-free combinations (\eg, ViM adding an energy term, or NNGuide~\cite{NNGuide-Park23} scaling logits) implicitly assume their constituent tests provide independent signals. As we demonstrate, this assumption fails: class-geometry scores are conditionally redundant with MSP under fusion, meaning a feature-based partner like fDBD~\cite{fDBD-Liu24} contributes only through its off-manifold component. Closest to our design is SIRC~\cite{SIRC-Xia22,SIRC-Xia24}, which combines the softmax score with a class-agnostic feature norm (a feature $L_1$~\cite{GradNorm-Huang21} or ViM residual norm). In contrast, our approach uses a \emph{class-conditional} off-manifold residual. Prior work has neither isolated the individual contributions of these design choices nor explained their inconsistent outcomes---a gap our factorial analysis addresses (Section~\ref{sec:analysis}).

\vspace{0.7em}
\noindent \textbf{Failure Prediction and Selective Classification.} Dedicated failure predictors like TrustScore~\cite{TrustScore-Jiang18}, ConfidNet~\cite{ConfidNet-Corbiere19}, and OpenMix~\cite{OpenMix-Zhu23} score misclassifications directly. ConfidNet and OpenMix train auxiliary heads, whereas TrustScore operates post-hoc and ranks predictions by agreement with class-conditional neighbors in feature space. Against this backdrop, none of these predictors isolates the signal that survives fusion with MSP.

\section{Preliminaries}
\label{sec:preliminaries}

\noindent \textbf{Selective Classification.} Let $f : \mathcal{X} \rightarrow \mathbb{R}^C$ be a neural network trained to classify inputs $x \in \mathcal{X}$ into one of $C$ classes, with predicted class $\hat{y} = \operatorname*{arg\,max}_c f_c(x)$. In the selective classification framework, the model is equipped with a confidence scoring function $s(x) \in \mathbb{R}$ and a threshold $\tau$: it outputs the prediction $\hat{y}$ if $s(x) \ge \tau$, and abstains otherwise. An ideal score ranks all correct predictions above all incorrect ones, so that errors are rejected first as the threshold $\tau$ increases.

\vspace{0.7em}
\noindent \textbf{The MSP Baseline and Feature Space.} We decompose $f$ into a feature extractor and a linear classifier head. Let $\phi(x) \in \mathbb{R}^{D}$ denote the penultimate feature vector; the logits are $z(x) = W\phi(x) + b$, with $W \in \mathbb{R}^{C \times D}$ and $b \in \mathbb{R}^{C}$ the classifier weights and biases. The predicted distribution is $p(x) = \mathrm{Softmax}(z(x))$, and Maximum Softmax Probability (MSP), the canonical failure-prediction baseline, defines confidence as the highest class probability, $s_{\mathrm{MSP}}(x) = \max_c p_c(x)$. Because $s_{\mathrm{MSP}}$ is derived solely from the $C$-dimensional logit vector $z(x)$, it collapses the $D$-dimensional feature geometry---a structural bottleneck we formally analyze in Section~\ref{sec:analysis}.

\vspace{0.7em}
\noindent \textbf{Evaluation Protocol: Isolating Ranking Quality.} The Risk--Coverage curve plots selective risk (the error rate on the accepted subset) against coverage as the threshold $\tau$ varies; AURC is its integral. Raw AURC is confounded by base accuracy---a more accurate model makes fewer errors and so scores a lower AURC regardless of ranking quality, and the oracle optimum $\mathrm{AURC}^{*}$ is fixed entirely by the base error rate. To isolate ranking quality we follow the fd-shifts protocol~\cite{Jaeger23} and adopt the Excess AURC, $\text{e-AURC}=\mathrm{AURC}-\mathrm{AURC}^{*}$~\cite{geifman2018bias} (reported $\times10^{3}$), as our primary metric: zero for a perfect predictor, lower is better, and comparable across backbones of differing accuracy. As secondary, gaming-resistant evidence we report AUGRC~\cite{traub2024overcoming} ($P(\text{accept}\cap\text{wrong})$, which abstaining on easy samples cannot inflate); AUROC serves only as a rank-sanity check.

\section{What Complements MSP? A Factorial Analysis of the Feature Residual}
\label{sec:analysis}

To build a detector that complements MSP without degrading baseline performance, we first isolate the exact information discarded by the classifier head, and then measure which of that discarded signal actually survives fusion. Existing feature-space methods entangle several design choices over a chosen feature, specifically the projection subspace (basis), the distance metric, and the reference population (conditioning). This entanglement makes it impossible to attribute their failures to any single factor. We deconstruct this space through a controlled factorial experiment (Figure~\ref{fig:factorial}). The orthogonal blind spot (Section~\ref{sec:analysis-blindspot}) and the conditional-redundancy principle (Section~\ref{sec:analysis-redundancy}) establish exactly why the feature residual is the correct geometric object. Following this, we determine how to evaluate it along the three fundamental axes of distance, basis, and conditioning.

\subsection{The Orthogonal Blind Spot of Linear Heads}
\label{sec:analysis-blindspot}
Every logit-based score factors through $z(x) = W\phi(x) + b$, and MSP further reduces $z$ to the scalar $\varphi(z) = \max_c \mathrm{Softmax}(z)_c$. By the chain rule, $\nabla_\phi s_{\mathrm{MSP}} = W^\top \nabla_z \varphi$ spans a single direction in $\mathbb{R}^D$. This \textbf{scalar collapse} leaves MSP first-order blind to the $(D-1)$-dimensional hyperplane orthogonal to that gradient---on ViT-B/16, 767 of the model's 768 feature dimensions. Since the gradient direction varies with the input, this statement is local. It therefore supplies only a necessary condition: a complementary signal must vary transverse to MSP's gradient. Which transverse component actually survives fusion is an empirical question, answered by the factorial analysis (Sections~\ref{sec:analysis-redundancy}--\ref{sec:analysis-conditioning}). This limitation persists even when the head has full column rank in feature space (here $D{=}768 < C{=}1000$, so $W$ can span all of $\mathbb{R}^D$): a full-rank head guarantees that the logit map discards nothing, but MSP's scalar collapse of the logits still does.

\subsection{The Conditional-Redundancy Principle}
\label{sec:analysis-redundancy}

\begin{figure*}[t]
  \centering
  \includegraphics[width=\textwidth]{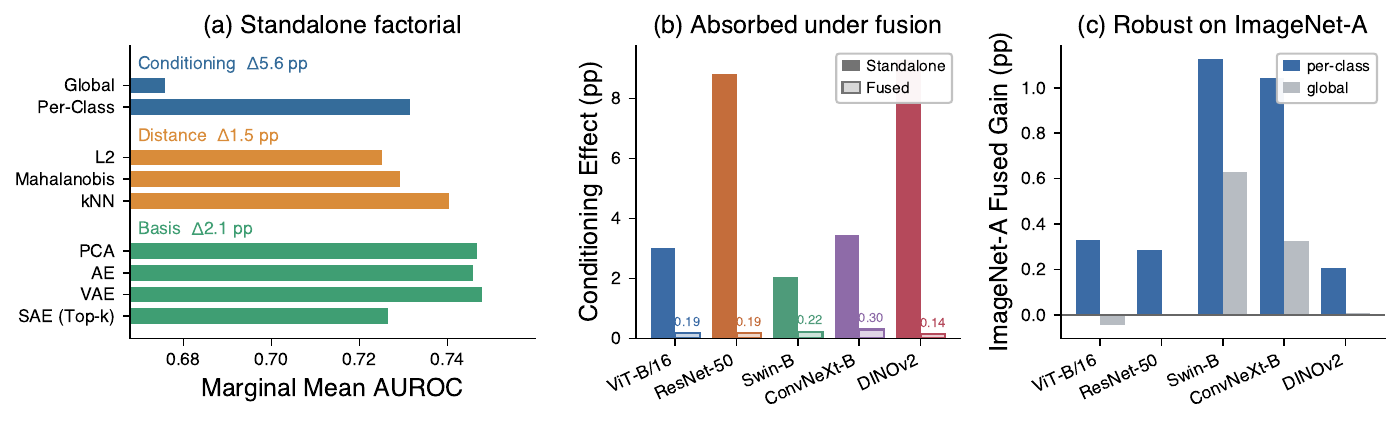}
  \caption{\textbf{The residual factorial and its fate under fusion.}
    \textbf{(a)}~Standalone main effects on the ViT-B/16 penultimate residual (validation AUROC): conditioning strictly dominates ($\Delta 5.6$ pp), while distance ($\Delta 1.5$ pp) and basis ($\Delta 2.1$ pp) are secondary. Marginals exclude the AE to prevent its Mahalanobis degeneracy from inflating the secondary axes. \textbf{(b)}~Fused with MSP, this large standalone advantage collapses to $\approx0.2$ pp across backbones (conditional redundancy). \textbf{(c)}~Per-class conditioning remains essential for robustness: on natural-adversarial ImageNet-A its fused gain stays positive on every backbone while global weakens and reverses (ViT-B/16).}
  \label{fig:factorial}
\end{figure*}

Cross-entropy training is expected to concentrate class-discriminative variation in the directions the head reads; if so, a distance computed in the full $D$-dimensional space (Mahalanobis~\cite{Lee18}, Deep kNN~\cite{Sun22}) is dominated by class-separability geometry---precisely what MSP already encodes. Conditioning is by far the dominant standalone axis ($\Delta 5.6$ pp AUROC; Figure~\ref{fig:factorial}a), dwarfing distance ($\Delta 1.5$ pp) and basis ($\Delta 2.1$ pp). But standalone performance is the wrong criterion for a detector deployed in fusion with MSP. Once fused, the conditioning advantage collapses to $\approx0.2$ pp on every backbone (Figure~\ref{fig:factorial}b): its variance is shared with MSP, which already encodes the class separability that conditioning exploits. We call this the conditional-redundancy principle: class-geometry signals are redundant with MSP under fusion. A detector escapes redundancy only by projecting away the principal subspace and scoring the residual $r(x)$.

\subsection{Distance: The Necessity of a Non-Parametric Estimator}
\label{sec:analysis-distance}

\begin{wrapfigure}{r}{0.45\textwidth}
  \vspace{-20pt}
  \centering
  \includegraphics[width=1.0\linewidth]{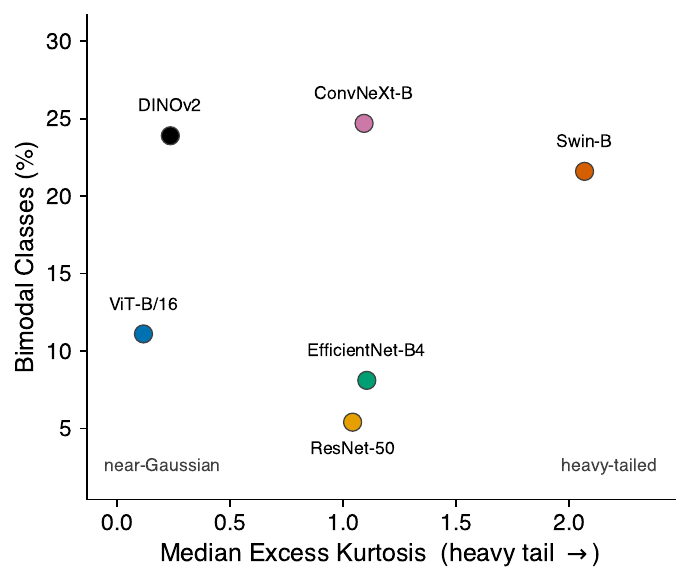}
  \caption{\textbf{Residual non-Gaussianity is backbone-dependent.} Median per-class excess kurtosis ($x$) vs.\ bimodal-class fraction ($y$), one point per backbone. The kurtosis axis separates the near-Gaussian and heavy-tailed regimes; bimodality does not. Per-class densities in Appendix~\ref{app:nongaussian}.}
  \label{fig:nongaussian}
  \vspace{-15pt}
\end{wrapfigure}

Beyond isolating $r(x)$, the scoring function must accurately reflect the residual's underlying distribution. Existing methods like ViM and Mahalanobis summarize the residual with parametric quantities (such as an $L_2$ norm or a shared-covariance Gaussian), implicitly assuming a zero-mean, near-isotropic density. The validity of this assumption is strongly backbone-dependent (Figure~\ref{fig:nongaussian}; Appendix~\ref{app:nongaussian}). For instance, the per-class residuals of ViT-B/16 and DINOv2 remain close to Gaussian (median excess kurtosis $0.12$ and $0.24$), whereas those of Swin-B and the CNNs are heavily leptokurtic (median $1.0$--$2.1$).

Consequently, no single parametric density transfers across these structural regimes. The standard shared-covariance assumption under-fits near-Gaussian residuals by ignoring class-specific variations. Conversely, modeling a full per-class covariance over-fits the heavy-tailed regimes (Swin-B and the CNNs), performing strictly worse than the shared-covariance baseline. The optimal parametric choice fluctuates with the backbone's geometry, and incorporating mixture components fails to close the performance gap (Appendix~\ref{app:density_ablation}). A non-parametric kNN estimator sidesteps this dilemma entirely. By measuring the distance to the $k$ nearest training-set neighbors, it reproduces the best parametric density on the near-Gaussian backbones (within ${\sim}0.01$ AUROC) and is the single best estimator where tails are heavy. No parametric family is uniformly best, whereas kNN is uniformly near-best.

\subsection{Basis: PCA and the Coordinate Problem}
\label{sec:analysis-basis}

\begin{wrapfigure}{r}{0.35\textwidth}
  \vspace{-15pt}
  \centering
  \includegraphics[width=1.0\linewidth]{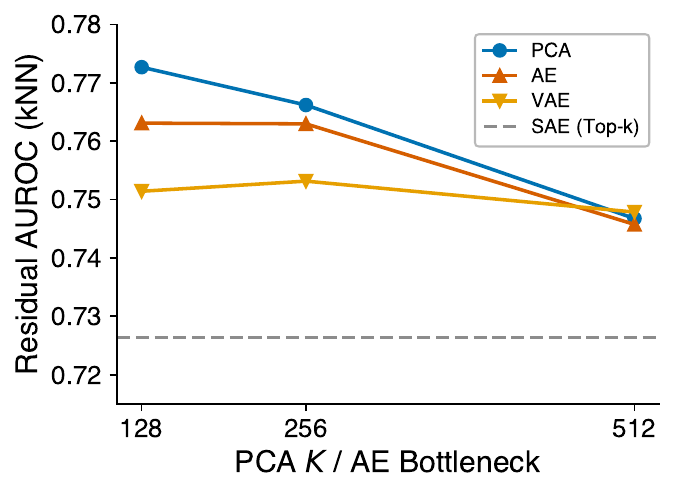}
  \caption{\textbf{Basis is inconsequential under kNN.}}
  \label{fig:basis}
  \vspace{-12pt}
\end{wrapfigure}

Under the deployed per-class kNN estimator, the choice of basis is nearly inconsequential. PCA, AE, and VAE~\cite{VAE-Kingma14} all perform within a $0.2$ pp margin, with the VAE's nominal $0.1$ pp advantage falling entirely within crossfit seed noise (Figure~\ref{fig:basis}). Because kNN evaluates local geometry directly, it is inherently coordinate-free, making the specific factorization of the residual largely irrelevant. The same holds for a cosine-kernel basis: replacing PCA with the cosine-kernel PCA of~\cite{NEURIPS2024_f2543511} moves CANARY's fused AUROC by at most $0.0007$ across four backbones (Appendix~\ref{app:kpca}). We therefore adopt PCA. Training-free, deterministic, and parameter-free, it matches the best performance at zero added computational cost. The single exception is the overcomplete TopK-SAE~\cite{Gao24}, trailing by ${\sim}2$ pp. Its dynamic, per-input atom selection projects same-class residuals into varying coordinate systems. This inconsistency dissolves the cohesive per-class clusters that kNN relies on to measure typicality (Appendix~\ref{app:basis_spectrum}).

\subsection{Conditioning: Class-Conditional vs. Global Typicality}
\label{sec:analysis-conditioning}
The last axis is the reference population. A global residual score asks whether a feature is atypical of the entire training set, whereas a class-conditional score asks whether it is atypical of its predicted class $\hat{y}$. For synthetic, near-isotropic corruptions a global score suffices, but structured natural shifts such as ImageNet-A and ImageNet-R induce class-dependent off-manifold variation, and under fusion the two scores diverge sharply (Figure~\ref{fig:factorial}c). There a global reference, blind to the predicted class, is actively misled exactly where per-class typicality still tracks error. Most of the standalone conditioning gap is redundant with MSP and collapses under fusion (Section~\ref{sec:analysis-redundancy}), but this robustness residual is not, and it is systematic. Across backbones the per-class residual variance shows a moderate positive correlation with the per-class failure rate (Spearman $0.44$--$0.64$; Appendix~\ref{app:perclass_var}). An error is best read as a violation of the predicted class's off-manifold structure, exposing the limits of global atypicality under shift.

\vspace{0.7em}
\noindent\textbf{Summary.} Each competing design fails under fusion for its own reason: a full-space distance duplicates MSP, a parametric norm mismodels the residual's distribution, and a global reference conditions on the wrong population. What survives all three tests---a non-parametric, class-conditional typicality on the orthogonal residual---is the blueprint for the method we detail in Section~\ref{sec:method}.

\section{Method}
\label{sec:method}
Our proposed method, \textbf{CANARY}, is the minimal instantiation of the blueprint distilled in Section~\ref{sec:analysis}, coupling the resulting residual score to MSP through a single validation-tuned weight. Let $\phi(x) \in \mathbb{R}^D$ denote the penultimate feature of a frozen classifier, and $\hat{y}(x)$ its predicted class. CANARY fits strictly on the ID-correct training samples, $\mathcal{D}^\checkmark = \{(x_i, y_i) \mid \hat{y}(x_i) = y_i\}$, to capture the typical geometry of correctly classified features.

\subsection{Low-Rank PCA Residual}

We compute PCA on the $\mathcal{D}^\checkmark$ features. Let $\mu \in \mathbb{R}^D$ be the mean and $V_K \in \mathbb{R}^{D \times K}$ the top-$K$ eigenvectors. The residual is the projection of the centered feature onto the orthogonal complement:
\begin{equation}
    r(x) = \bigl(I - V_K V_K^\top\bigr)\bigl(\phi(x) - \mu\bigr).
\end{equation}
This projection isolates the feature variation left unexplained by the class-separating directions, the redundant component MSP already captures. We fix $K=512$. The residual's $K$-dependent variation is precisely that redundant class geometry. Fusion therefore absorbs it, rendering the fused score insensitive to $K$ on every backbone, even where the residual's overall gain is large (Appendix~\ref{app:k-dependence}).

\subsection{Per-Class kNN Distance}

For each class $c$, we group the $\mathcal{D}^\checkmark$ residuals into a reference set $\mathcal{R}_c = \bigl\{r(x_i) \mid (x_i, y_i) \in \mathcal{D}^\checkmark,\ y_i = c\bigr\}$. At inference, we score $r(x)$ by its Euclidean distance $d(x) = \bigl\lVert r(x) - r^{(k)} \bigr\rVert_2$ to its $k$-th nearest neighbor $r^{(k)}$ ($k=100$) in the predicted class's reference set $\mathcal{R}_{\hat{y}(x)}$. These reference sets merely store the training residuals, requiring no covariance fitting or auxiliary training.

\subsection{Percentile Calibration}

Because the kNN distance $d(x)$ is unbounded and backbone-dependent, while MSP operates on an unrelated scale, the two signals must be placed on a common footing prior to fusion. Passing a score through the empirical CDF of its reference distribution performs the probability integral transform, mapping the score to a $p$-value that is uniform on $[0,1]$ for typical inputs. Let $\mathcal{V}$ denote the set of correctly classified validation inputs:
\begin{equation}
    s_r(x) = 1 - \hat{F}\bigl(d(x)\bigr), \qquad \hat{F}(t) = \frac{1}{|\mathcal{V}|} \sum_{x_j \in \mathcal{V}} \mathds{1}\bigl[d(x_j) \le t\bigr],
\end{equation}
where $\mathds{1}[\cdot]$ denotes the indicator function (1 if the condition holds, 0 otherwise). A higher $s_r$ thus marks a more typical, likely-correct input.

We calibrate MSP identically against its own validation-correct CDF to obtain $s_m$. Both signals are now $p$-values computed against a fixed validation reference, reflecting precisely what a deployed system can calculate without access to test labels. These standardized $p$-values provide well-calibrated inputs for a Fisher-inspired log-linear combination~\cite{fisher1970statistical}.
  
\subsection{Weighted-Fisher Fusion}

Percentile calibration yields two $p$-values per input: $s_m(x)$ from MSP and $s_r(x)$ from the residual. A Fisher-style combination multiplies the two $p$-values (equivalently, sums their logs). We apply a weighted formulation in the logarithmic domain:
\begin{equation}
    S(x;\, w) = w \ln s_m(x) + (1 - w) \ln s_r(x).
\end{equation}
At $w = \tfrac{1}{2}$, this recovers the canonical Fisher combination ($s_m(x) s_r(x)$), which assumes equally informative signals. However, our signals are strongly asymmetric. By the conditional-redundancy principle, MSP already captures the residual's class-relevant signal. Fusion absorbs this redundancy, isolating only the residual's unique off-manifold component. MSP thus supplies the dominant evidence, prioritized through the validation-tuned weight $w$.

We tune $w$ via cross-fitting on the clean validation set. Specifically, we split the set in half, maximizing AUROC on one and evaluating on the other. We repeat this across five random seeds and average the results. Because the fusion relies on a single scalar weight, this low-capacity formulation mitigates over-adaptation to the validation splits. The optimal weight $w^\star$ is then frozen for all subsequent shift evaluations. We treat this as a log-linear pooling rule rather than a hypothesis test: since $s_m$ and $s_r$ are dependent (Section~\ref{sec:analysis-redundancy}), we neither invoke nor rely on the test's independence-based null distribution; $w$ is tuned empirically and the score is used only for ranking.

\section{Experiments}
\label{sec:experiments}

\subsection{Setup}

\noindent \textbf{Datasets.} We study failure prediction on ImageNet-1k~\cite{ImageNet-Deng09,ILSVRC-Russakovsky15}, where the objective is to robustly separate a model's correct predictions from its own misclassifications under distribution shift. ImageNet-1k validation (50k) serves as the clean, in-distribution reference. We evaluate robustness across three distribution-shift families: ImageNet-C~\cite{ImageNetC-Hendrycks19} (15 corruptions $\times$ 5 severities), ImageNet-A~\cite{ImageNetA-Hendrycks21} (natural adversarials), and ImageNet-R~\cite{ImageNetR-Hendrycks21b} (renditions). ObjectNet~\cite{ObjectNet-Barbu19} (an independently collected, controlled real-world covariate shift) is held out as a further generalization test. To test transfer to a different label space, we additionally evaluate on CIFAR-100~\cite{CIFAR-Krizhevsky09} and its corrupted variant CIFAR-100-C~\cite{ImageNetC-Hendrycks19}.

\vspace{0.7em}
\noindent \textbf{Backbones.} To establish architectural generality, we evaluate five main backbones: ViT-B/16~\cite{ViT-Dosovitskiy21} as the primary architecture, along with ResNet-50~\cite{ResNet-He16}, Swin-B~\cite{Swin-Liu21}, ConvNeXt-B~\cite{ConvNeXt-Liu22}, and EfficientNet-B4~\cite{EfficientNet-Tan19}. We further probe representational generality using the self-supervised DINOv2~\cite{DINOv2-Oquab24}. To isolate the drivers of CANARY's gain within a single architecture family, we include three additional ViT configurations: one differently fine-tuned ViT-B/16 that lowers baseline MSP quality while keeping the architecture, size, and pretraining data constant; and two scaled variants (ViT-S/16 and ViT-L/16) that shift MSP headroom through capacity alone, with ViT-L/16 approaching the accuracy ceiling. Finally, the CIFAR-100 experiments employ the standard OpenOOD v1.5~\cite{OpenOODv15-Zhang23} ResNet-18 and a WideResNet-28-10~\cite{WideResNet-Zagoruyko16} trained with a standard recipe.

For every backbone, we fix CANARY's basis rank $K{=}512$ ($K = D-128$ for the narrower 384-d ViT-S/16 and 512-d CIFAR ResNet-18) and neighbor count $k{=}100$, and cross-fit the single fusion weight $w$ strictly on that backbone's own clean validation split. Full details on the pretrained checkpoints and the per-backbone optimum $w^\star$ are provided in Appendix~\ref{app:models}.

\vspace{0.7em}
\noindent \textbf{Baselines and metrics.} CANARY is compared against 19 post-hoc baselines spanning logit-based, activation-shaping, feature-based, and softmax-combination families. The primary metric is e-AURC, the accuracy-corrected excess area under the risk--coverage curve, which ensures fair comparability across backbones of differing base accuracies; AUGRC and AUROC serve as secondary metrics.

\subsection{Experimental Results}
\label{sec:experimental-results}

\begin{table}[t]
\centering
\caption{\textbf{Failure prediction across three backbones spanning the MSP-headroom range (low$\rightarrow$high).} Mean over the 77 distribution-shift cells (ImageNet-C 15$\times$5, ImageNet-A, ImageNet-R). e-AURC and AUGRC are reported $\times 10^3$. Best in \textbf{bold}, second and third \underline{underlined}.}
\label{tab:main_results}
\resizebox{\textwidth}{!}{
\begin{tabular}{lccccccccc}
\toprule
\multirow{2}{*}{\textbf{Method}} & \multicolumn{3}{c}{\textbf{ViT-B/16}} & \multicolumn{3}{c}{\textbf{ConvNeXt-B}} &
\multicolumn{3}{c}{\textbf{EfficientNet-B4}} \\
\cmidrule(lr){2-4} \cmidrule(lr){5-7} \cmidrule(lr){8-10}
& \textbf{e-AURC} $\downarrow$ & \textbf{AUGRC} $\downarrow$ & \textbf{AUROC} $\uparrow$ & \textbf{e-AURC} $\downarrow$ & \textbf{AUGRC} $\downarrow$ &
\textbf{AUROC} $\uparrow$ & \textbf{e-AURC} $\downarrow$ & \textbf{AUGRC} $\downarrow$ & \textbf{AUROC} $\uparrow$ \\
\midrule
\multicolumn{10}{l}{\textit{Logit-based}} \\
\midrule
MSP~\cite{hendrycks2017baseline} & \underline{58.2} & \underline{98.2} & \underline{0.8537} & \underline{61.4} & \underline{96.1} & \underline{0.8531} &
\underline{87.5} & \underline{155.3} & \underline{0.8318} \\
MaxLogit~\cite{Hendrycks2022ScalingOD} & 84.7 & 108.6 & 0.8016 & 89.7 & 104.8 & 0.8083 & 123.0 & 165.2 & 0.7854 \\
Energy~\cite{liu2020energy} & 97.6 & 114.4 & 0.7739 & 112.4 & 114.4 & 0.7601 & 182.4 & 189.1 & 0.6746 \\
Doctor~\cite{Granese21} & \underline{58.9} & \underline{98.7} & \underline{0.8514} & 61.7 & 96.3 & 0.8520 & 88.8 & 155.9
& 0.8286 \\
GEN~\cite{GEN-Liu23} & 85.3 & 109.0 & 0.7993 & 79.0 & 103.4 & 0.8149 & 103.6 & 161.4 & 0.8027 \\
\midrule
\multicolumn{10}{l}{\textit{Activation-shaping}} \\
\midrule
ASH~\cite{Djurisic23} & 159.4 & 138.5 & 0.6568 & 108.6 & 114.5 & 0.7593 & 236.0 & 210.4 & 0.5777 \\
ReAct~\cite{Sun21} & 99.4 & 115.1 & 0.7707 & 91.6 & 108.5 & 0.7901 & 175.1 & 191.6 & 0.6637 \\
DICE~\cite{Sun22b} & 104.7 & 117.3 & 0.7601 & 136.0 & 124.7 & 0.7089 & 252.4 & 217.3 & 0.5455 \\
\midrule
\multicolumn{10}{l}{\textit{Feature-based}} \\
\midrule
ViM~\cite{ViM-Wang22} & 128.8 & 128.5 & 0.7069 & 94.0 & 111.1 & 0.7761 & 123.3 & 171.5 & 0.7538 \\
Deep kNN~\cite{Sun22} & 270.5 & 173.6 & 0.4939 & 247.1 & 160.5 & 0.5251 & 193.8 & 186.6 & 0.6837 \\
MDS~\cite{Lee18} & 320.1 & 189.1 & 0.4231 & 77.6 & 104.9 & 0.8070 & 109.7 & 166.9 & 0.7756 \\
MDS++~\cite{MDSpp-Mueller25} & 302.9 & 182.1 & 0.4559 & 80.3 & 106.0 & 0.8021 & 104.1 & 164.9 & 0.7857 \\
RMDS~\cite{RMDS-Ren21} & 260.2 & 171.8 & 0.5039 & 70.5 & 102.1 & 0.8218 & 95.8 & 161.3 & 0.8021 \\
RMDS++~\cite{MDSpp-Mueller25} & 163.7 & 138.8 & 0.6559 & 74.7 & 104.2 & 0.8121 & 104.6 & 165.6 & 0.7826 \\
NNGuide~\cite{NNGuide-Park23} & 89.0 & 110.9 & 0.7901 & 78.5 & 104.7 & 0.8085 & 108.9 & 165.7 & 0.7808 \\
fDBD~\cite{fDBD-Liu24} & 76.8 & 105.8 & 0.8150 & 67.5 & 100.4 & 0.8300 & 95.1 & 159.6 & 0.8106 \\
TrustScore~\cite{TrustScore-Jiang18} & 69.3 & 104.0 & 0.8262 & 62.7 & 99.7 & 0.8357 & 89.6 & 160.6 & 0.8073 \\
\midrule
\multicolumn{10}{l}{\textit{Softmax-combination}} \\
\midrule
SIRC (L1)~\cite{SIRC-Xia22} & 65.4 & 102.1 & 0.8362 & 62.7 & 96.4 & 0.8515 & 89.7 & 156.0 & 0.8285 \\
SIRC (Residual)~\cite{SIRC-Xia22} & 60.1 & 99.7 & 0.8471 & \underline{59.3} & \underline{96.0} & \underline{0.8533} & \underline{83.1} & \underline{155.1} & \underline{0.8317} \\
\midrule
\rowcolor{gray!15} \textbf{CANARY (ours)} & \textbf{57.8} & \textbf{98.0} & \textbf{0.8545} & \textbf{56.9} & \textbf{95.1} & \textbf{0.8576} & \textbf{78.0}
& \textbf{153.3} & \textbf{0.8406} \\
\bottomrule
\end{tabular}
}
\end{table}

\noindent\textbf{CANARY ranks first across backbones and metrics.} On the main ImageNet-1k backbones, CANARY outperforms all 19 baselines on e-AURC, AUGRC, and AUROC, the sole exception being a within-seed-noise near-tie with SIRC (Residual) on ResNet-50 (Appendix~\ref{app:sirc-backbone}). Table~\ref{tab:main_results} highlights three backbones spanning the MSP-headroom range, with full-panel comparisons deferred to Appendix~\ref{app:full-result}. The performance margin over MSP follows the headroom law: the gap peaks where the baseline MSP is weakest and narrows toward parity as MSP improves. Consequently, absolute e-AURC reductions vary by over an order of magnitude, from 9.5 ($\times 10^3$) on the low-MSP EfficientNet-B4 to a tighter $0.4$ on the high-MSP ViT-B/16. The standalone feature-based detectors trail by a wide margin, even the strongest of them (TrustScore or fDBD), by $0.02$--$0.03$ AUROC; the logit scores (MaxLogit, Energy, GEN) trail throughout. The two SIRC variants isolate the signal question within a fixed combination function: SIRC (Residual) (an off-manifold residual norm) beats SIRC (L1) (a global feature norm) on every backbone, independent evidence that the complementary signal is off-manifold rather than global magnitude---the choice CANARY sharpens further with a per-class residual, surpassing both.

\begin{table*}[htbp]
\centering
\caption{\textbf{Reliability across the five main backbones (385 shift cells).} ``Top-1 Cells'' counts cells where the method is the single best of the 20 detectors; ``$\ge$ MSP Cells'' counts cells where the method matches or exceeds MSP; Max Drop is its largest regression relative to MSP. Only detectors that top at least one cell are listed. CANARY dominates the win-count and is the only detector with a bounded worst case; SIRC (Residual)'s top-1 cells are concentrated on ResNet-50, where it is a near-tie with CANARY within seed noise~(Appendix~\ref{app:sirc-backbone}).}
\label{tab:reliability}
\vspace{0.5em} 
\resizebox{0.85\textwidth}{!}{ %
\begin{tabular}{lcccccc} 
\toprule
\multirow{2}{*}{\textbf{Method}} & \multicolumn{3}{c}{\textbf{e-AURC} ($\times 10^3$)} & \multicolumn{3}{c}{\textbf{AUROC}} \\
\cmidrule(lr){2-4} \cmidrule(lr){5-7}
& \textbf{Top-1 Cells} & \textbf{$\ge$ MSP Cells} & \textbf{Max Drop} & \textbf{Top-1 Cells} & \textbf{$\ge$ MSP Cells} & \textbf{Max Drop} \\
\midrule
MSP    & 3  & --- & ---      & 3  & --- & ---       \\
\midrule
SIRC (Residual) & 35 & 289 & $-$15.0  & 56 & 215  & $-$0.0394 \\
fDBD & 16 & 53  & $-$25.9  & 15 & 40  & $-$0.0662 \\
TrustScore   & 16 & 130  & $-$46.1  & 13 & 17  & $-$0.1334 \\
Doctor & 4  & 64  & $-$17.0  & 4  & 59  & $-$0.0318 \\
RMDS & 2  & 34  & $-$352.8 & 4  & 15  & $-$0.4378 \\
ASH & 1  & 1   & $-$249.7 & 0  & 1   & $-$0.3340 \\
GEN & 0  & 12  & $-$45.8  & 1  & 11  & $-$0.1154 \\
\midrule
\textbf{CANARY (ours)} & \textbf{308} & \textbf{382} & \textbf{$-$0.1} & \textbf{289} & \textbf{379} & \textbf{$-$0.0002} \\
\bottomrule
\end{tabular}
}
\end{table*}

\vspace{0.7em}
\noindent \textbf{CANARY is the only detector never worse than MSP.} By degrading to MSP when the residual provides no new information, CANARY matches or exceeds the baseline in 382 of 385 shift cells on e-AURC (379 on AUROC). Table~\ref{tab:reliability} compares this robustness against all baselines that achieve the top rank in at least one cell. While CANARY is the single best detector in 308 of 385 cells by e-AURC, no competitor matches its reliability. Its nearest rival here, the combination baseline SIRC (Residual), tops 35 cells but is not never-worse (worst-cell $-0.039$ AUROC, $-15.0$ e-AURC); the standalone feature detectors fare worse still---fDBD, RMDS, and CANARY's closest sibling TrustScore collapse far below MSP on backbones whose geometry does not suit them, with worst-case e-AURC regressions of $25.9$, $352.8$, and $46.1$ ($\times 10^3$), respectively, against CANARY's $0.1$. By anchoring the residual to MSP, CANARY keeps its worst-case regression bounded to the seed-noise floor, gracefully trading marginal peak performance on a few cells rather than collapsing on any.

\vspace{0.7em}
\noindent \textbf{The headroom law across architectures.} The dependence of the performance margin on MSP quality is a consistent cross-architecture trend rather than a per-model artifact. Figure~\ref{fig:headroom-law} plots the fusion AUROC gain against each backbone's clean-validation MSP AUROC. The backbones align along a single negatively sloped trend, and the optimal fusion weight $w^\star$ tracks this available headroom, decreasing monotonically from $0.95$ on high-MSP backbones to $0.70$ on the weaker EfficientNet-B4. This trend holds across both synthetic (ImageNet-C) and natural (ImageNet-A/R) shifts, extends to the self-supervised DINOv2, and is mirrored on the held-out ObjectNet (Appendix~\ref{app:headroom-objectnet}).

Crucially, this law holds even within a single architecture. By fixing the ViT-B/16 structure and ImageNet-21k pretraining while varying only the fine-tuning strategy (\texttt{augreg} with MSP $0.8783$ vs.\ \texttt{augreg2} with $0.8651$), the variant with the weaker baseline MSP achieves substantially larger gains ($+0.42$ vs.\ $+0.08$ pp AUROC across shifts, and $+1.04$ vs.\ $\approx$0 on ObjectNet). This isolates MSP quality as the primary driver of the performance margin, free from architectural confounds.

\subsection{Generality Across Regimes}
\label{sec:result-generality}

\noindent \textbf{CIFAR-100: the never-worse property transfers to a different label space.} We evaluate CANARY on CIFAR-100 to test whether its never-worse property holds in a genuinely different label space (100 classes, disjoint from ImageNet). To this end, we use two CIFAR-native models (the standard OpenOOD v1.5 ResNet-18 and a WideResNet-28-10 trained with a standard recipe) that are entirely independent of ImageNet on both the data and pretraining axes. CANARY stays at or above MSP on 136 of 150 cells across the two backbones, with the 14 exceptions on WRN-28-10 all falling within a $-0.0004$ AUROC margin that represents effective ties rather than true losses. The mean gains are small because both are CNN backbones whose off-manifold residual is nearly uninformative (standalone residual AUROC $\approx 0.55$--$0.57$ vs.\ $\approx 0.75$ on ViT features). Consequently, the fusion weight sits at the MSP anchor ($w^\star = 0.9$--$0.95$), consistent with the headroom law. While the boundary-based detector fDBD achieves a higher shift-mean on these cleanly separated features, it lacks the never-worse property across the ImageNet backbones, dropping by up to 0.066 AUROC relative to MSP (Table~\ref{tab:reliability}). CANARY trades a marginally lower peak for an empirically bounded worst case across every setting we evaluate.

\begin{table*}[htbp]
\centering
\caption{\textbf{Generality across label spaces and representations.} We evaluate CANARY beyond the standard ImageNet setting on a different label space (CIFAR-100) and a self-supervised backbone (DINOv2). ``$\ge$ MSP Cells'' denotes the number of shift cells where CANARY matches or outperforms MSP. $\dagger$ indicates that CANARY outperforms MSP on every cell. AUROC values represent the mean across all shift cells.}
\label{tab:generality}
\resizebox{0.95\textwidth}{!}{ 
\begin{tabular}{ll c ccc c}
\toprule
\multirow{2}{*}{\textbf{Axis}} & \multirow{2}{*}{\textbf{Backbone}} & \multirow{2}{*}{\textbf{MSP}} & \multicolumn{3}{c}{\textbf{CANARY (Ours)}} & \multirow{2}{*}{\textbf{Best Baseline} (AUROC$\uparrow$)} \\
\cmidrule(lr){4-6}
& & & \textbf{AUROC$\uparrow$} & \textbf{$\ge$ MSP Cells} & \textbf{Max Drop} & \\
\midrule
\multirow{2}{*}{Label Space} 
& CIFAR-100 ResNet-18 & 0.7915 & 0.7959 & 75/75 & $+0.0005^\dagger$ & fDBD (0.8007) \\
& CIFAR-100 WRN-28-10 & 0.7871 & 0.7882 & 61/75 & $-$0.0004         & fDBD (0.7996) \\
\midrule
Representation 
& DINOv2 (self-supervised)  & 0.8519 & 0.8538 & 77/77 & $+0.0010^\dagger$ & SIRC (L1) (0.8550) \\
\bottomrule
\end{tabular}
}
\end{table*}

\vspace{0.7em}
\noindent \textbf{DINOv2: the never-worse property extends to a self-supervised representation.} On DINOv2, a representation learned entirely without class labels, CANARY still improves over MSP, never regresses on any shift cell, and its gain grows monotonically with corruption severity. The fact that the off-manifold residual remains informative on a representation trained without supervision indicates that the mechanism is intrinsic to the feature geometry, rather than an artifact of label-supervised training. Although SIRC (L1) attains a marginally higher shift-mean on DINOv2 ($0.8550$ vs.\ $0.8538$), it trails CANARY on all five main backbones.

\vspace{0.7em}
\noindent \textbf{Model scale: the never-worse property holds across the ViT family.} We further vary model capacity along the ViT hierarchy, specifically evaluating ViT-S/16 and ViT-L/16, the smaller and larger siblings of the primary ViT-B/16. On the smaller ViT-S, the residual continues to complement MSP, and CANARY maintains strict baseline dominance across every shift cell. On the larger ViT-L, where the baseline MSP approaches the accuracy ceiling, the residual is less consistent in adding novel information: while the mean gain persists, CANARY marginally trails MSP on 11 of 77 cells, though all such cases fall within seed noise (worst case $-0.0005$ AUROC). Across scales, CANARY never falls meaningfully below MSP, while the number of cells at or above MSP decreases as available headroom shrinks ($77 \to 74 \to 66$; Appendix~\ref{app:vit-scale}).

\subsection{Computational Cost}

CANARY adds negligible overhead to existing MSP pipelines. Fitting is training- and backpropagation-free: given the extracted features, building the PCA basis and per-class reference set requires $\approx$1.9~TFLOP, and the resulting reference set occupies just $\approx$0.6~GB for ImageNet-1k on ViT-B/16. At inference, a 768$\to$256 projection and a single $L_2$ distance add merely $\approx$1.3~MFLOP per query. On an NVIDIA GeForce RTX 4080, this translates to an overhead of $\approx$50\,$\mu$s per unbatched query, averaging under 2\,$\mu$s batched.

\section{Conclusion}
\label{sec:conclusion}

This work elucidates the mechanisms governing MSP-complementary failure prediction: the conditional-redundancy principle shows that the class-relevant geometry of a frozen feature space is absorbed by MSP under fusion, so only the orthogonal, off-manifold component contributes novel information. CANARY operationalizes this principle, measuring class-conditional non-parametric typicality on the PCA residual and fusing it with MSP via a single validation-tuned weight. Evaluated under a strictly deployment-realistic protocol without retraining or OOD data exposure, CANARY delivers its largest gains where MSP is weakest, closely following the headroom law, while degrading gracefully toward MSP where the baseline is already strong. Crucially, it remains empirically never-worse, not regressing meaningfully below the baseline MSP across hundreds of distribution shifts, diverse backbone architectures, and even a different label space.

\vspace{0.7em}
\noindent \textbf{Limitations.} We identify two primary limitations. First, while our findings are robustly validated across diverse shift sets and backbones and grounded in a geometric rationale, they remain empirical. We have yet to provide a rigorous mathematical formalization regarding the joint distribution of the fused scores. The headroom law, in particular, is an empirical regularity established over the nine backbones and shift families studied here rather than an analytical derivation, and its exact functional form may differ beyond this regime. Second, CANARY is structurally blind to purely on-manifold confident failures, where a shifted input is genuinely typical of an incorrect class and leaves no off-manifold trace. However, this represents only a subset of hard natural shifts. On ImageNet-A, for instance, CANARY still improves upon MSP because many challenging samples retain an off-manifold component; only the strictly on-manifold remainder lies beyond its reach.

\vspace{0.7em}
\noindent \textbf{Future work.} Subsequent research should aim to formalize the conditions under which the empirical never-worse property holds, and to derive an analytical bound for the headroom law, currently established as an empirical forecast. Methodologically, closing the on-manifold gap requires integrating external signals (\eg, boundary-distance or raw-input heuristics). Fusing these with CANARY's off-manifold typicality offers a promising path toward comprehensive failure prediction without surrendering its never-worse reliability. Finally, expanding CANARY to dense prediction tasks (\eg, per-pixel typicality), adversarial settings, and high-stakes domains (\eg, medical imaging) remains an important direction.

\vspace{0.7em}
\noindent \textbf{Acknowledgments}
This work was supported by the National Research Foundation of Korea (NRF) grant (No.~RS-2026-25490405) and by the Institute of Information \& Communications Technology Planning \& Evaluation (IITP) grant (No.~RS-2024-00428780), both funded by the Korea government (MSIT). 

\newpage

\bibliographystyle{splncs04}
\bibliography{main}

@String(IJCV  = {Int. J. Comput. Vis.})

@String(CVPR  = {IEEE Conf. Comput. Vis. Pattern Recog.})

@String(ICCV  = {Int. Conf. Comput. Vis.})

@String(ECCV  = {Eur. Conf. Comput. Vis.})

@String(NeurIPS = {Adv. Neural Inform. Process. Syst.})

@String(ICML  = {Int. Conf. Mach. Learn.})

@String(ICLR  = {Int. Conf. Learn. Represent.})

@String(ACCV  = {Asian Conf. Comput. Vis.})

@String(BMVC  = {Brit. Mach. Vis. Conf.})

@String(TMLR  = {Trans. Mach. Learn Res.})

@inproceedings{hendrycks2017baseline,
  author    = {Hendrycks, Dan and Gimpel, Kevin},
  title     = {A Baseline for Detecting Misclassified and Out-of-Distribution Examples in Neural Networks},
  booktitle = ICLR,
  year      = {2017}
}

@inproceedings{Hendrycks2022ScalingOD,
  author    = {Hendrycks, Dan and Basart, Steven and Mazeika, Mantas and Zou, Andy and Kwon, Joseph and Mostajabi, Mohammadreza and Steinhardt, Jacob and Song, Dawn},
  title     = {Scaling Out-of-Distribution Detection for Real-World Settings},
  booktitle = ICML,
  pages     = {8759--8773},
  year      = {2022}
}

@inproceedings{liu2020energy,
  author    = {Liu, Weitang and Wang, Xiaoyun and Owens, John and Li, Yixuan},
  title     = {Energy-based Out-of-distribution Detection},
  booktitle = NeurIPS,
  pages     = {21464--21475},
  year      = {2020}
}

@inproceedings{Granese21,
  author    = {Granese, Federica and Romanelli, Marco and Gorla, Daniele and Palamidessi, Catuscia and Piantanida, Pablo},
  title     = {{DOCTOR}: A Simple Method for Detecting Misclassification Errors},
  booktitle = NeurIPS,
  pages     = {5669--5681},
  year      = {2021}
}

@inproceedings{GEN-Liu23,
  author    = {Liu, Xixi and Lochman, Yaroslava and Zach, Christopher},
  title     = {{GEN}: Pushing the Limits of Softmax-Based Out-of-Distribution Detection},
  booktitle = CVPR,
  pages     = {23946--23955},
  year      = {2023}
}

@inproceedings{Sun21,
  author    = {Sun, Yiyou and Guo, Chuan and Li, Yixuan},
  title     = {{ReAct}: Out-of-distribution Detection With Rectified Activations},
  booktitle = NeurIPS,
  pages     = {144--157},
  year      = {2021}
}

@inproceedings{Djurisic23,
  author    = {Djurisic, Andrija and Bozanic, Nebojsa and Ashok, Arjun and Liu, Rosanne},
  title     = {Extremely Simple Activation Shaping for Out-of-Distribution Detection},
  booktitle = ICLR,
  year      = {2023}
}

@inproceedings{Sun22b,
  author    = {Sun, Yiyou and Li, Yixuan},
  title     = {{DICE}: Leveraging Sparsification for Out-of-Distribution Detection},
  booktitle = ECCV,
  pages     = {691--708},
  year      = {2022}
}

@inproceedings{Lee18,
  author    = {Lee, Kimin and Lee, Kibok and Lee, Honglak and Shin, Jinwoo},
  title     = {A Simple Unified Framework for Detecting Out-of-Distribution Samples and Adversarial Attacks},
  booktitle = NeurIPS,
  year      = {2018}
}

@inproceedings{Sun22,
  author    = {Sun, Yiyou and Ming, Yifei and Zhu, Xiaojin and Li, Yixuan},
  title     = {Out-of-Distribution Detection with Deep Nearest Neighbors},
  booktitle = ICML,
  pages     = {20827--20840},
  year      = {2022}
}

@inproceedings{ViM-Wang22,
  author    = {Wang, Haoqi and Li, Zhizhong and Feng, Litong and Zhang, Wayne},
  title     = {{ViM}: Out-of-Distribution With Virtual-Logit Matching},
  booktitle = CVPR,
  pages     = {4921--4930},
  year      = {2022}
}

@inproceedings{NNGuide-Park23,
  author    = {Park, Jaewoo and Jung, Yoon Gyo and Teoh, Andrew Beng Jin},
  title     = {Nearest Neighbor Guidance for Out-of-Distribution Detection},
  booktitle = ICCV,
  pages     = {1686--1695},
  year      = {2023}
}

@inproceedings{fDBD-Liu24,
  author    = {Liu, Litian and Qin, Yao},
  title     = {Fast Decision Boundary based Out-of-Distribution Detector},
  booktitle = ICML,
  pages     = {31728--31746},
  year      = {2024}
}

@inproceedings{NEURIPS2024_f2543511,
  author    = {Fang, Kun and Tao, Qinghua and Lv, Kexin and He, Mingzhen and Huang, Xiaolin and Yang, Jie},
  title     = {Kernel {PCA} for Out-of-Distribution Detection},
  booktitle = NeurIPS,
  pages     = {134317--134344},
  year      = {2024}
}

@inproceedings{guan2023revisit,
  author    = {Guan, Xiaoyuan and Liu, Zhouwu and Zheng, Wei-Shi and Zhou, Yuren and Wang, Ruixuan},
  title     = {Revisit {PCA}-based Technique for Out-of-Distribution Detection},
  booktitle = ICCV,
  pages     = {19431--19439},
  year      = {2023}
}

@article{RMDS-Ren21,
  author    = {Ren, Jie and Fort, Stanislav and Liu, Jeremiah and Roy, Abhijit Guha and Padhy, Shreyas and Lakshminarayanan, Balaji},
  title     = {A Simple Fix to {Mahalanobis} Distance for Improving Near-{OOD} Detection},
  journal   = {arXiv preprint arXiv:2106.09022},
  year      = {2021}
}

@inproceedings{MDSpp-Mueller25,
  author    = {M{\"u}ller, Maximilian and Hein, Matthias},
  title     = {{Mahalanobis++}: Improving {OOD} Detection via Feature Normalization},
  booktitle = ICML,
  pages     = {45151--45184},
  year      = {2025}
}

@inproceedings{GradNorm-Huang21,
  author    = {Huang, Rui and Geng, Andrew and Li, Yixuan},
  title     = {On the Importance of Gradients for Detecting Distributional Shifts in the Wild},
  booktitle = NeurIPS,
  pages     = {677--689},
  year      = {2021}
}

@inproceedings{SIRC-Xia22,
  author    = {Xia, Guoxuan and Bouganis, Christos-Savvas},
  title     = {Augmenting Softmax Information for Selective Classification with Out-of-Distribution Data},
  booktitle = ACCV,
  pages     = {664--680},
  year      = {2023}
}

@article{SIRC-Xia24,
  author    = {Xia, Guoxuan and Bouganis, Christos-Savvas},
  title     = {Augmenting the Softmax with Additional Confidence Scores for Improved Selective Classification with Out-of-Distribution Data},
  journal   = IJCV,
  volume    = {132},
  number    = {9},
  pages     = {3714--3752},
  year      = {2024}
}

@inproceedings{TrustScore-Jiang18,
  author    = {Jiang, Heinrich and Kim, Been and Guan, Melody and Gupta, Maya},
  title     = {To Trust Or Not To Trust A Classifier},
  booktitle = NeurIPS,
  year      = {2018}
}

@inproceedings{ConfidNet-Corbiere19,
  author    = {Corbi{\`e}re, Charles and Thome, Nicolas and Bar-Hen, Avner and Cord, Matthieu and P{\'e}rez, Patrick},
  title     = {Addressing Failure Prediction by Learning Model Confidence},
  booktitle = NeurIPS,
  year      = {2019}
}

@inproceedings{OpenMix-Zhu23,
  author    = {Zhu, Fei and Cheng, Zhen and Zhang, Xu-Yao and Liu, Cheng-Lin},
  title     = {{OpenMix}: Exploring Outlier Samples for Misclassification Detection},
  booktitle = CVPR,
  pages     = {12074--12083},
  year      = {2023}
}

@inproceedings{Jaeger23,
  author    = {Jaeger, Paul F. and L{\"u}th, Carsten T. and Klein, Lukas and Bungert, Till J.},
  title     = {A Call to Reflect on Evaluation Practices for Failure Detection in Image Classification},
  booktitle = ICLR,
  year      = {2023}
}

@inproceedings{traub2024overcoming,
  author    = {Traub, Jeremias and Bungert, Till J. and L{\"u}th, Carsten T. and Baumgartner, Michael and Maier-Hein, Klaus H. and Maier-Hein, Lena and J{\"a}ger, Paul F.},
  title     = {Overcoming Common Flaws in the Evaluation of Selective Classification Systems},
  booktitle = NeurIPS,
  pages     = {2323--2347},
  year      = {2024}
}

@inproceedings{geifman2018bias,
  author    = {Geifman, Yonatan and Uziel, Guy and El-Yaniv, Ran},
  title     = {Bias-Reduced Uncertainty Estimation for Deep Neural Classifiers},
  booktitle = ICLR,
  year      = {2019}
}

@book{fisher1970statistical,
  author    = {Fisher, Ronald A.},
  title     = {Statistical Methods for Research Workers},
  publisher = {Oliver and Boyd},
  address   = {Edinburgh},
  year      = {1925}
}

@inproceedings{ViT-Dosovitskiy21,
  author    = {Dosovitskiy, Alexey and Beyer, Lucas and Kolesnikov, Alexander and Weissenborn, Dirk and Zhai, Xiaohua and Unterthiner, Thomas and Dehghani, Mostafa and Minderer, Matthias and Heigold, Georg and Gelly, Sylvain and Uszkoreit, Jakob and Houlsby, Neil},
  title     = {An Image is Worth 16x16 Words: Transformers for Image Recognition at Scale},
  booktitle = ICLR,
  year      = {2021}
}

@inproceedings{ResNet-He16,
  author    = {He, Kaiming and Zhang, Xiangyu and Ren, Shaoqing and Sun, Jian},
  title     = {Deep Residual Learning for Image Recognition},
  booktitle = CVPROLD,
  pages     = {770--778},
  year      = {2016}
}

@inproceedings{Swin-Liu21,
  author    = {Liu, Ze and Lin, Yutong and Cao, Yue and Hu, Han and Wei, Yixuan and Zhang, Zheng and Lin, Stephen and Guo, Baining},
  title     = {{Swin Transformer}: Hierarchical Vision Transformer Using Shifted Windows},
  booktitle = ICCV,
  pages     = {10012--10022},
  year      = {2021}
}

@inproceedings{ConvNeXt-Liu22,
  author    = {Liu, Zhuang and Mao, Hanzi and Wu, Chao-Yuan and Feichtenhofer, Christoph and Darrell, Trevor and Xie, Saining},
  title     = {A {ConvNet} for the 2020s},
  booktitle = CVPR,
  pages     = {11976--11986},
  year      = {2022}
}

@inproceedings{EfficientNet-Tan19,
  author    = {Tan, Mingxing and Le, Quoc},
  title     = {{EfficientNet}: Rethinking Model Scaling for Convolutional Neural Networks},
  booktitle = ICML,
  pages     = {6105--6114},
  year      = {2019}
}

@article{DINOv2-Oquab24,
  author    = {Oquab, Maxime and Darcet, Timoth{\'e}e and Moutakanni, Th{\'e}o and Vo, Huy V. and Szafraniec, Marc and Khalidov, Vasil and Fernandez, Pierre and Haziza, Daniel and Massa, Francisco and El-Nouby, Alaaeldin and Assran, Mido and Ballas, Nicolas and Galuba, Wojciech and Howes, Russell and Huang, Po-Yao and Li, Shang-Wen and Misra, Ishan and Rabbat, Michael and Sharma, Vasu and Synnaeve, Gabriel and Xu, Hu and J{\'e}gou, Herv{\'e} and Mairal, Julien and Labatut, Patrick and Joulin, Armand and Bojanowski, Piotr},
  title     = {{DINOv2}: Learning Robust Visual Features without Supervision},
  journal   = TMLR,
  year      = {2024}
}

@inproceedings{WideResNet-Zagoruyko16,
  author    = {Zagoruyko, Sergey and Komodakis, Nikos},
  title     = {Wide Residual Networks},
  booktitle = BMVC,
  pages     = {87.1--87.12},
  year      = {2016}
}

@inproceedings{ImageNet-Deng09,
  author    = {Deng, Jia and Dong, Wei and Socher, Richard and Li, Li-Jia and Li, Kai and Fei-Fei, Li},
  title     = {{ImageNet}: A Large-Scale Hierarchical Image Database},
  booktitle = CVPROLD,
  pages     = {248--255},
  year      = {2009}
}

@article{ILSVRC-Russakovsky15,
  author    = {Russakovsky, Olga and Deng, Jia and Su, Hao and Krause, Jonathan and Satheesh, Sanjeev and Ma, Sean and Huang, Zhiheng and Karpathy, Andrej and Khosla, Aditya and Bernstein, Michael and Berg, Alexander C. and Fei-Fei, Li},
  title     = {{ImageNet} Large Scale Visual Recognition Challenge},
  journal   = IJCV,
  volume    = {115},
  number    = {3},
  pages     = {211--252},
  year      = {2015}
}

@inproceedings{ImageNetC-Hendrycks19,
  author    = {Hendrycks, Dan and Dietterich, Thomas},
  title     = {Benchmarking Neural Network Robustness to Common Corruptions and Perturbations},
  booktitle = ICLR,
  year      = {2019}
}

@inproceedings{ImageNetA-Hendrycks21,
  author    = {Hendrycks, Dan and Zhao, Kevin and Basart, Steven and Steinhardt, Jacob and Song, Dawn},
  title     = {Natural Adversarial Examples},
  booktitle = CVPR,
  pages     = {15262--15271},
  year      = {2021}
}

@inproceedings{ImageNetR-Hendrycks21b,
  author    = {Hendrycks, Dan and Basart, Steven and Mu, Norman and Kadavath, Saurav and Wang, Frank and Dorundo, Evan and Desai, Rahul and Zhu, Tyler and Parajuli, Samyak and Guo, Mike and Song, Dawn and Steinhardt, Jacob and Gilmer, Justin},
  title     = {The Many Faces of Robustness: A Critical Analysis of Out-of-Distribution Generalization},
  booktitle = ICCV,
  pages     = {8340--8349},
  year      = {2021}
}

@inproceedings{ObjectNet-Barbu19,
  author    = {Barbu, Andrei and Mayo, David and Alverio, Julian and Luo, William and Wang, Christopher and Gutfreund, Dan and Tenenbaum, Josh and Katz, Boris},
  title     = {{ObjectNet}: A Large-Scale Bias-Controlled Dataset for Pushing the Limits of Object Recognition Models},
  booktitle = NeurIPS,
  year      = {2019}
}

@techreport{CIFAR-Krizhevsky09,
  author      = {Krizhevsky, Alex},
  title       = {Learning Multiple Layers of Features from Tiny Images},
  institution = {University of Toronto},
  year        = {2009}
}

@article{OpenOODv15-Zhang23,
  author    = {Zhang, Jingyang and Yang, Jingkang and Wang, Pengyun and Wang, Haoqi and Lin, Yueqian and Zhang, Haoran and Sun, Yiyou and Du, Xuefeng and Li, Yixuan and Liu, Ziwei and Chen, Yiran and Li, Hai},
  title     = {{OpenOOD} v1.5: Enhanced Benchmark for Out-of-Distribution Detection},
  journal   = DMLR,
  volume    = {2},
  number    = {3},
  pages     = {1--32},
  year      = {2024}
}

@inproceedings{VAE-Kingma14,
  author    = {Kingma, Diederik P. and Welling, Max},
  title     = {Auto-Encoding Variational {Bayes}},
  booktitle = ICLR,
  year      = {2014}
}

@inproceedings{Gao24,
  author    = {Gao, Leo and {Dupr{\'e} la Tour}, Tom and Tillman, Henk and Goh, Gabriel and Troll, Rajan and Radford, Alec and Sutskever, Ilya and Leike, Jan and Wu, Jeffrey},
  title     = {Scaling and Evaluating Sparse Autoencoders},
  booktitle = ICLR,
  year      = {2025}
}

\ifsupp
\clearpage

\appendix

\begin{center}
  {\Large\bf{Supplementary Material for\\ ``A Canary in the Neural Mine: Catching an Image Classifier's Silent, Confident Failures''}\par}
\end{center}
\medskip

\section{Non-Gaussianity of the Residual Across All Backbones}
\label{app:nongaussian}

Section~\ref{sec:analysis} summarized the non-Gaussianity regimes (Figure~\ref{fig:nongaussian}); this section provides the full per-backbone picture. On the same per-class PCA residual that the detector deploys, we measure, per class, heavy-tailedness (pooled per-dimension excess kurtosis $\kappa$; 0 = Gaussian) and multi-modality (Hartigan's dip test on the dominant within-class direction; bimodal if $p < 0.05$).

\vspace{0.7em}
\noindent \textbf{Per-backbone densities.} For each backbone, Figure~\ref{fig:nongaussian-full} plots the standardized residual density against a fitted Gaussian on a logarithmic axis. To provide a transparent view of the within-backbone spread, we pair the most non-Gaussian class with the median (typical) class. Swin-B and the CNNs (ConvNeXt-B, ResNet-50, and EfficientNet-B4) exhibit heavy tails even at the median. The high-MSP ViT-B/16 remains near-Gaussian throughout. While the extreme class of the self-supervised DINOv2 is explicitly bimodal, its typical class remains near-Gaussian, aligning with its minority 24\% bimodal fraction.

\begin{figure}[h]
  \centering
  \includegraphics[width=0.7\textwidth]{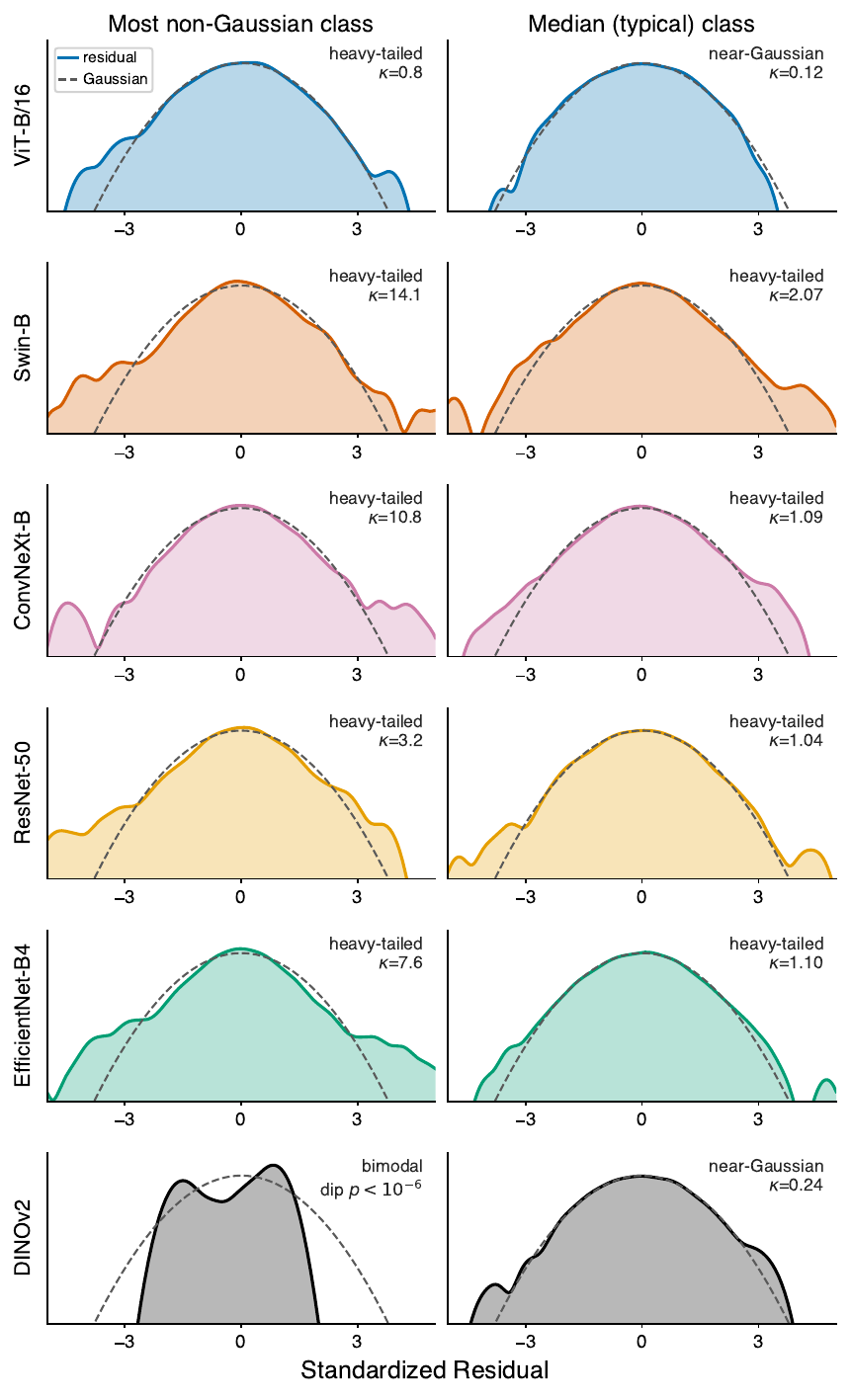}
  \caption{\textbf{Per-backbone residual non-Gaussianity.} We pair each backbone's most non-Gaussian class (left) with its median (typical) class (right). Each panel plots the standardized residual density (solid) against a fitted Gaussian (dashed) on a logarithmic $y$-axis, annotated with its regime and excess kurtosis $\kappa$.}
  \label{fig:nongaussian-full}
\end{figure}

\vspace{0.7em}
\noindent \textbf{Regime summary.} Figure~\ref{fig:nongaussian} in the main text maps the median per-class excess kurtosis against the bimodal-class fraction, separating the backbones into distinct tail regimes. Swin-B and the CNNs occupy the heavy-tailed corner, while ViT-B/16 and DINOv2 are near-Gaussian. Table~\ref{tab:nongaussian_stats} summarizes the measured statistics, highlighting that DINOv2 carries the highest bimodal fraction despite being predominantly near-Gaussian.

\begin{table}[h]
  \centering
  \caption{Per-class residual statistics and geometric regimes across backbones.}
  \label{tab:nongaussian_stats}
  \resizebox{\linewidth}{!}{%
  \begin{tabular}{lccc}
    \toprule
    \textbf{Backbone} & \textbf{Bimodal \% (Dip $p < 0.05$)} & \textbf{Median Excess Kurtosis} & \textbf{Dominant Regime} \\
    \midrule
    ViT-B/16 & 10.8\% & 0.12 & Near-Gaussian \\
    Swin-B & 22.0\% & 2.07 & Heavy-tailed \\
    ConvNeXt-B & 24.0\% & 1.09 & Heavy-tailed \\
    ResNet-50 & 5.4\% & 1.04 & Heavy-tailed \\
    EfficientNet-B4 & 8.1\% & 1.10 & Heavy-tailed \\
    DINOv2 & 24.4\% & 0.24 & Near-Gaussian, most bimodal classes \\
    \bottomrule
  \end{tabular}%
  }
\end{table}

\clearpage

\section{Density-Model Ablation}
\label{app:density_ablation}

This section expands on Section~\ref{sec:analysis-distance}. The tail heaviness of the per-class residual is backbone-dependent, so no single parametric family transfers across architectures. A non-parametric kNN, by contrast, adapts robustly to each.

\vspace{0.7em}
\noindent \textbf{The best parametric density flips with the regime.} Table~\ref{tab:density_ablation} reports the standalone residual-only AUROC for different estimators, using the deployed per-class conditioning and $k=100$ for the kNN.

\begin{table}[h]
  \centering
  \caption{Standalone AUROC across parametric and non-parametric estimators.}
  \label{tab:density_ablation}
  \resizebox{1.0\linewidth}{!}{
  \begin{tabular}{l c c c c c c}
    \toprule
    \textbf{Estimator} & \textbf{ViT-B/16}  & \textbf{ConvNeXt-B} & \textbf{Swin-B} & \textbf{ResNet-50} & \textbf{EfficientNet-B4} & \textbf{DINOv2} \\
    \midrule
    Median excess kurtosis & 0.12 & 1.09 & 2.07 & 1.04 & 1.10 & 0.24 \\
    \midrule
    Mahalanobis (shared $\Sigma$) & 0.728 & 0.766 & 0.754 & 0.712 & 0.755 & 0.707 \\
    Diagonal GMM (best $M \leq 32$) & 0.740 & 0.764 & 0.752 & 0.715 & 0.757 & 0.725 \\
    Full per-class $\Sigma$ Gaussian & 0.748 & 0.746 & 0.740 & 0.703 & 0.749 & 0.735 \\
    \midrule
    \rowcolor{gray!15}
    \textbf{Per-class kNN ($k=100$)} & 0.747 & 0.771 & 0.760 & 0.722 & 0.758 & 0.729 \\
    \bottomrule
  \end{tabular}
  }
\end{table}

On the near-Gaussian backbones (ViT-B/16, DINOv2), the full per-class covariance Gaussian is the strongest parametric density and closely matches kNN, while the shared-covariance Mahalanobis of prior work trails it by failing to capture class-specific covariances. This ordering inverts on the heavy-tailed backbones (Swin-B and the CNNs), where the full covariance over-fits the tails and drops below the shared-covariance baseline. The kNN estimator is uniformly competitive---matching the best parametric density on the near-Gaussian backbones (within ${\sim}0.01$ AUROC) and leading on the heavy-tailed ones.

\vspace{0.7em}
\noindent \textbf{Multimodality is not the primary lever.} Sweeping the number of mixture components in a per-class diagonal GMM does not reach kNN performance on any backbone, and full-covariance GMMs with $M > 1$ consistently degrade the result. This holds even for DINOv2, the most bimodal backbone (24\% of classes flagged by dip test). Because its typical class remains near-Gaussian (median excess kurtosis 0.24; median dip $p=0.82$), fitting extra modes merely over-fits the unimodal majority. The residual's departure from Gaussianity is primarily a heavy-tail phenomenon, not a mode-count issue.

\vspace{0.7em}
\noindent \textbf{kNN robustness.} The kNN estimator requires no regime-specific tuning or covariance estimation. It consistently matches or exceeds the shared-covariance Mahalanobis baseline, providing an AUROC gain of 0.003 to 0.022. Against the best parametric density on each backbone, it leads on every heavy-tailed backbone and stays within 0.006 AUROC on the near-Gaussian ones.

\clearpage

\section{Why the TopK SAE basis underperforms}
\label{app:basis_spectrum}

Section~\ref{sec:analysis-basis} evaluates the choice of basis under the deployed per-class kNN estimator. This section provides a finer geometric perspective underlying those results.

The SAE residual does not occupy a fixed subspace. Instead, its overcomplete code selects a different set of active neurons per input. Consequently, same-class residuals lack a shared coordinate system, loosening the per-class clusters that kNN relies on. Because the code is trained strictly to reconstruct, its leftover is dominated by hard-to-reconstruct, instance-specific details.

Empirically, reconstruction quality and failure-prediction performance do not align. PCA at a bottleneck of $K=256$ leaves 27.8\% of the feature variance unexplained (0.2775). The overcomplete TopK SAE (latent dimension 6144, sparsity 32) leaves a smaller 26.2\% (0.2623) unexplained. However, under the per-class kNN, the SAE yields an AUROC of 0.7264, falling strictly behind PCA's 0.7662---a ${\sim}4$ pp gap at this reconstruction-matched $K=256$; against the deployed $K=512$ PCA (AUROC 0.747) the SAE still trails by the ${\sim}2$ pp quoted in Section~\ref{sec:analysis-basis}. A smaller reconstruction residual thus does not buy a better failure-prediction score. This confirms that the cause is fundamentally geometric rather than a matter of residual size. Any fixed basis recovers the off-manifold geometry equally well, making PCA the natural, training-free choice.

\begin{figure}[h]
  \centering
  \includegraphics[width=0.5\textwidth]{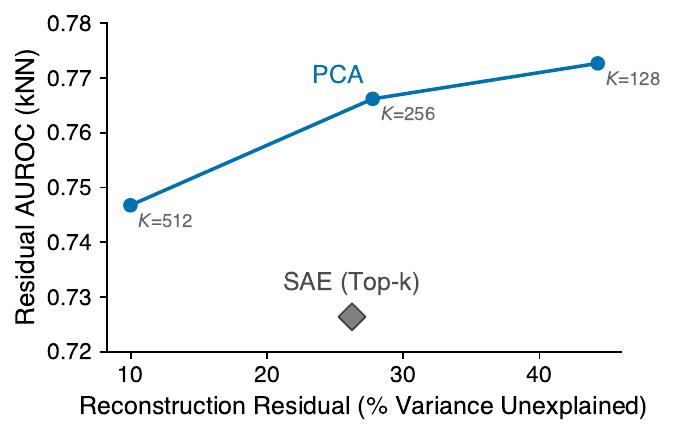}
  \caption{\textbf{The SAE underperformance is geometric.} On ViT-B/16, despite reconstructing slightly better than PCA, the overcomplete TopK SAE scores worse under the per-class kNN.}
  \label{fig:sae_deficit}
  \vspace{-15pt}
\end{figure}

\section{Per-Class Residual Variance Predicts Failure}
\label{app:perclass_var}

The advantage of the per-class conditioning discussed in Section~\ref{sec:analysis-conditioning} is systematic. Specifically, a class whose in-distribution (ID) correct residuals are more dispersed is inherently more failure-prone. For each class $c$, we measure the per-class residual total variance over its ID-correct training residuals:
\begin{equation}
    \operatorname{tr}(\Sigma_c) = \frac{1}{|\mathcal{C}_c|} \sum_{i \in \mathcal{C}_c} \lVert r_i - \mu_c \rVert^2,
\end{equation}
where $\mathcal{C}_c$ is the set of ID-correct training samples for class $c$, and $\mu_c$ is their residual mean. Alongside this, we calculate the per-class validation failure rate, defined as $\mathrm{fail}_c = 1 - \mathrm{acc}(\text{val} \mid \text{true class } c)$. We then compute the Spearman rank correlation between the residual variances $\{\operatorname{tr}(\Sigma_c)\}$ and the failure rates $\{\mathrm{fail}_c\}$ across all 1,000 classes for each backbone.

\begin{table}[h]
  \centering
  \caption{Spearman rank correlation between per-class residual variance and failure rate.}
  \label{tab:variance_failure}
  \begin{tabular}{lc}
    \toprule
    \textbf{Backbone} & \textbf{Spearman $\rho$ (Variance vs.\ Failure)} \\
    \midrule
    ViT-B/16 & 0.55 \\
    ConvNeXt-B & 0.64 \\
    Swin-B & 0.61 \\
    ResNet-50 & 0.49 \\
    EfficientNet-B4 & 0.61 \\
    DINOv2 & 0.44 \\
    \bottomrule
  \end{tabular}
\end{table}

The correlation is consistently positive and moderate on every backbone, ranging from 0.44 to 0.64. Consequently, the per-class residual spread systematically tracks where errors are most likely to occur. A global reference inherently discards this crucial class-specific failure information by pooling all classes into a single distribution.

\section{Robustness to the Basis Rank $K$}
\label{app:k-dependence}

\noindent \textbf{CANARY is $K$-insensitive across backbones.} Sweeping $K$ at matched fractions of the feature dimension $D$ and evaluating across all shift families, the fused AUROC of CANARY varies by at most $0.0021$ over $K/D \in [0, 0.9]$---even on the 2048-d ResNet-50, where $K=512$ leaves the bottom three-quarters of the spectrum as the residual (Figure~\ref{fig:kpca}a). This is not a vanishing-gain artifact: on the large-gain Swin-B (Section~\ref{sec:experimental-results}), the score moves by under $0.001$ across $K \in \{0, \ldots, 512\}$.

\begin{figure}[h]
\centering
\includegraphics[width=\textwidth]{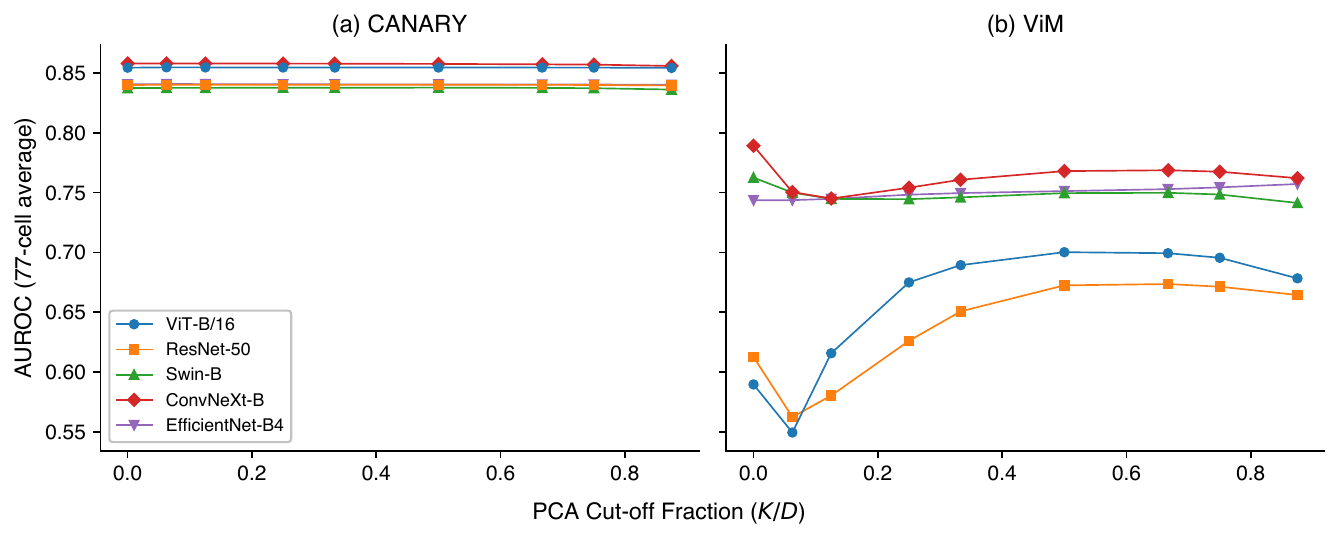}
\caption{\textbf{Robustness to the PCA basis rank $K$.} (a) On the five main backbones, the CANARY score is robust to the PCA rank $K$ (varying by $\le 0.0021$). (b) A global ViM-style residual norm is strongly $K$-dependent. The $y$-axis is the mean AUROC over the full 77 shift cells.}
\label{fig:kpca}
\end{figure}

\vspace{0.7em}
\noindent \textbf{Why fusion absorbs the $K$-dependence.} A small $K$ retains the class-mean directions in the residual---the class-separating axes the classifier head reads, and thus already encoded by MSP (Section~\ref{sec:analysis-redundancy}). Conditioned on MSP, these carry no additional information, so fusion discards them regardless of $K$. This robustness is specific to CANARY's per-class \emph{position} statistic: a global ViM-style norm depends on the subspace dimension and swings by $0.014$--$0.151$ across the same backbones (Figure~\ref{fig:kpca}b).

We fix $K=512$ at no performance cost. Because the projection is orthogonal, the trailing $(D{-}K)$-dimensional residual preserves the full-$D$ distances, so the reference set is stored and searched in that reduced space (256-d for ViT-B). Narrower backbones ($D \le 512$) instead set $K = D-128$, retaining a minimal 128-d residual (ViT-S/16, CIFAR ResNet-18).

\section{Backbone Checkpoints and Hyperparameters}
\label{app:models}

Every backbone is a standard pretrained checkpoint, evaluated frozen. Table~\ref{tab:backbone_hyperparams} details each checkpoint (a \texttt{timm} tag for the ImageNet-1k backbones), its clean top-1 accuracy, the MSP validation AUROC that anchors the headroom law, its penultimate feature dimension $D$, the basis rank $K$, and the cross-fitted fusion weight $w^\star$. The neighbor count ($k=100$) and the basis rank are not tuned per backbone but fixed by a single rule, $K=\min(512,\,D-128)$: cap the basis at $512$ principal components while always retaining at least a $128$-dimensional residual. 

\begin{table}[h]
\centering
\setlength{\tabcolsep}{10pt}
\caption{\textbf{Backbone checkpoints and CANARY hyperparameters.} $K=512$ for every backbone with $D\ge640$, and $K=256/384$ on the 384-d ViT-S~/ 512-d CIFAR ResNet-18 (each leaving a 128-d residual).}
\label{tab:backbone_hyperparams}
\resizebox{\textwidth}{!}{
\begin{tabular}{llccccc}
\toprule
\textbf{Backbone} & \textbf{Checkpoint} & \textbf{Top-1 (\%)} & \textbf{MSP AUROC} & \boldmath{$D$} & \boldmath{$K$} & \boldmath{$w^\star$} \\
\midrule
\multicolumn{7}{l}{\textbf{ImageNet-1k}} \\
ViT-B/16 (main)                & \texttt{vit\_base\_patch16\_224.augreg\_in21k\_ft\_in1k}  & 84.5 & 0.8783 & 768  & 512 & 0.95 \\
ViT-B/16 (within-arch control) & \texttt{vit\_base\_patch16\_224.augreg2\_in21k\_ft\_in1k} & 85.1 & 0.8651 & 768  & 512 & 0.80 \\
ViT-S/16                       & \texttt{vit\_small\_patch16\_224.augreg\_in21k\_ft\_in1k} & 81.4 & 0.8764 & 384  & 256 & 0.95 \\
ViT-L/16                       & \texttt{vit\_large\_patch16\_224.augreg\_in21k\_ft\_in1k} & 85.8 & 0.8837 & 1024 & 512 & 0.95 \\
ResNet-50                      & \texttt{resnet50.a1\_in1k}                                & 80.4 & 0.8757 & 2048 & 512 & 0.95 \\
Swin-B                         & \texttt{swin\_base\_patch4\_window7\_224.ms\_in22k\_ft\_in1k} & 85.3 & 0.8607 & 1024 & 512 & 0.80 \\
ConvNeXt-B                     & \texttt{convnext\_base.fb\_in22k\_ft\_in1k}               & 85.8 & 0.8714 & 1024 & 512 & 0.80 \\
EfficientNet-B4                & \texttt{efficientnet\_b4.ra2\_in1k}                       & 82.7 & 0.8482 & 1792 & 512 & 0.70 \\
DINOv2                         & \texttt{vit\_base\_patch14\_dinov2.lvd142m}               & 83.8 & 0.8737 & 768  & 512 & 0.95 \\
\midrule
\multicolumn{7}{l}{\textbf{CIFAR-100}} \\
ResNet-18                      & Pre-trained on OpenOOD v1.5    & 77.3 & 0.8752 & 512 & 384 & 0.90 \\
WideResNet-28-10               & Standard training recipe  & 81.7 & 0.8852 & 640 & 512 & 0.95 \\
\bottomrule
\end{tabular}
}
\end{table}

\newpage

\section{Full Experimental Results}
\label{app:full-result}

\subsection{Main Backbones}

Tables~\ref{tab:full-vitb}--\ref{tab:full-efficientnetb4} report all 20 detectors on the five main backbones. ImageNet-C is summarized by its four corruption groups~\cite{ImageNetC-Hendrycks19}; ImageNet-A/R are the full sets. e-AURC is reported $\times10^3$. Best detector is in \textbf{bold} and the second/third \underline{underlined}. Each table's bottom row ($\Delta$ vs.\ MSP) is CANARY${-}$MSP ($\uparrow$ = CANARY improves over MSP, $\downarrow$ = worse).

\begin{table}[htbp]
\centering
\caption{\textbf{ViT-B/16.}}
\label{tab:full-vitb}
\resizebox{\textwidth}{!}{
\begin{tabular}{l cc cc cc cc cc cc}
\toprule
\multirow{3}{*}{\textbf{Method}} & \multicolumn{8}{c}{\textbf{ImageNet-C}} & \multicolumn{2}{c}{\multirow{2}{*}{\textbf{ImageNet-A}}} & \multicolumn{2}{c}{\multirow{2}{*}{\textbf{ImageNet-R}}} \\
\cmidrule(lr){2-9}
 & \multicolumn{2}{c}{\textbf{Noise}} & \multicolumn{2}{c}{\textbf{Blur}} & \multicolumn{2}{c}{\textbf{Weather}} & \multicolumn{2}{c}{\textbf{Digital}} & \multicolumn{2}{c}{} & \multicolumn{2}{c}{} \\
\cmidrule(lr){2-3}\cmidrule(lr){4-5}\cmidrule(lr){6-7}\cmidrule(lr){8-9}\cmidrule(lr){10-11}\cmidrule(lr){12-13}
 & \textbf{e-AURC}$\downarrow$ & \textbf{AUROC}$\uparrow$ & \textbf{e-AURC}$\downarrow$ & \textbf{AUROC}$\uparrow$ & \textbf{e-AURC}$\downarrow$ & \textbf{AUROC}$\uparrow$ & \textbf{e-AURC}$\downarrow$ & \textbf{AUROC}$\uparrow$ & \textbf{e-AURC}$\downarrow$ & \textbf{AUROC}$\uparrow$ & \textbf{e-AURC}$\downarrow$ & \textbf{AUROC}$\uparrow$ \\
\midrule
\multicolumn{13}{l}{\textit{Logit-based}} \\
\midrule
MSP & \underline{53.9} & \underline{0.8575} & \underline{64.4} & \underline{0.8531} & \underline{50.5} & \underline{0.8598} & \underline{51.2} & \underline{0.8551} & 235.7 & 0.6842 & 114.6 & 0.8272 \\
MaxLogit & 79.6 & 0.8074 & 89.3 & 0.8091 & 78.2 & 0.8038 & 79.9 & 0.7944 & 265.6 & 0.6443 & 117.0 & 0.8204 \\
Energy & 92.7 & 0.7804 & 102.6 & 0.7823 & 90.2 & 0.7769 & 93.2 & 0.7642 & 280.7 & 0.6142 & 126.7 & 0.8039 \\
Doctor & 54.7 & 0.8550 & \underline{65.1} & \underline{0.8509} & 51.1 & 0.8577 & 52.1 & 0.8524 & 236.5 & 0.6821 & \underline{113.9} & \underline{0.8283} \\
GEN & 82.4 & 0.8016 & 89.3 & 0.8083 & 78.6 & 0.8016 & 80.0 & 0.7925 & 261.7 & 0.6503 & 116.8 & 0.8227 \\
\midrule
\multicolumn{13}{l}{\textit{Activation-shaping}} \\
\midrule
ASH & 144.6 & 0.6857 & 169.2 & 0.6683 & 150.0 & 0.6575 & 158.3 & 0.6279 & 339.6 & 0.5252 & 216.5 & 0.6850 \\
ReAct & 96.8 & 0.7733 & 104.0 & 0.7799 & 92.0 & 0.7734 & 93.3 & 0.7636 & 281.3 & 0.6125 & 131.3 & 0.7975 \\
DICE & 99.3 & 0.7685 & 110.2 & 0.7682 & 96.5 & 0.7635 & 100.4 & 0.7493 & 290.8 & 0.5952 & 141.5 & 0.7825 \\
\midrule
\multicolumn{13}{l}{\textit{Feature-based}} \\
\midrule
ViM & 127.1 & 0.7075 & 140.6 & 0.7090 & 119.0 & 0.7093 & 116.6 & 0.7070 & 288.8 & 0.5938 & 211.7 & 0.6935 \\
Deep kNN & 268.5 & 0.4921 & 312.8 & 0.4770 & 264.8 & 0.4877 & 232.8 & 0.5103 & 359.7 & 0.4809 & 333.0 & 0.5341 \\
MDS & 352.0 & 0.3708 & 359.5 & 0.4097 & 298.4 & 0.4427 & 268.2 & 0.4575 & 369.1 & 0.4677 & 478.4 & 0.3530 \\
MDS++ & 319.1 & 0.4331 & 338.0 & 0.4415 & 286.3 & 0.4667 & 264.5 & 0.4745 & 358.5 & 0.4932 & 403.7 & 0.4587 \\
RMDS & 288.4 & 0.4600 & 289.2 & 0.4995 & 236.4 & 0.5259 & 220.0 & 0.5250 & 351.6 & 0.4897 & 445.8 & 0.4038 \\
RMDS++ & 168.6 & 0.6444 & 164.8 & 0.6788 & 148.1 & 0.6675 & 157.7 & 0.6412 & 354.3 & 0.4905 & 304.2 & 0.5959 \\
NNGuide & 85.2 & 0.7919 & 93.7 & 0.7993 & 82.0 & 0.7929 & 83.1 & 0.7838 & 277.2 & 0.6272 & 122.5 & 0.8085 \\
fDBD & 74.8 & 0.8142 & 82.9 & 0.8195 & 70.6 & 0.8178 & 70.0 & 0.8122 & \textbf{230.8} & \textbf{0.6992} & \textbf{94.1} & \textbf{0.8559} \\
TrustScore & 66.7 & 0.8264 & 76.4 & 0.8264 & 60.7 & 0.8327 & 60.3 & 0.8294 & 241.4 & 0.6776 & 147.9 & 0.7758 \\
\midrule
\multicolumn{13}{l}{\textit{Softmax-combination}} \\
\midrule
SIRC (L1) & \underline{54.5} & \underline{0.8554} & 70.3 & 0.8373 & 55.2 & 0.8457 & 67.7 & 0.8197 & 236.2 & 0.6806 & 115.9 & 0.8224 \\
SIRC (Residual) & 61.1 & 0.8366 & 65.9 & 0.8479 & \underline{50.9} & \underline{0.8578} & \underline{51.5} & \underline{0.8524} & \underline{233.4} & \underline{0.6880} & 114.7 & 0.8243 \\
\midrule
\rowcolor{gray!15} \textbf{CANARY (ours)} & \textbf{53.8} & \textbf{0.8577} & \textbf{64.1} & \textbf{0.8536} & \textbf{50.1} & \textbf{0.8606} & \textbf{50.5} & \textbf{0.8565} & \underline{234.5} & \underline{0.6865} & \underline{113.5} & \underline{0.8289} \\
{\scriptsize ($\Delta$ vs.\ MSP)} & {\scriptsize 0.2$\uparrow$} & {\scriptsize 0.0002$\uparrow$} & {\scriptsize 0.3$\uparrow$} & {\scriptsize 0.0005$\uparrow$} & {\scriptsize 0.4$\uparrow$} & {\scriptsize 0.0009$\uparrow$} & {\scriptsize 0.7$\uparrow$} & {\scriptsize 0.0014$\uparrow$} & {\scriptsize 1.3$\uparrow$} & {\scriptsize 0.0023$\uparrow$} & {\scriptsize 1.1$\uparrow$} & {\scriptsize 0.0017$\uparrow$} \\
\bottomrule
\end{tabular}
}
\end{table}

\begin{table}[htbp]
\centering
\caption{\textbf{ViT model scale: MSP vs.\ CANARY.} ViT-S/16, ViT-B/16, and ViT-L/16 on ImageNet-C by corruption group (cell-means) and ImageNet-A/R. e-AURC $\times 10^3$ ($\downarrow$), AUROC ($\uparrow$); better of MSP/CANARY in \textbf{bold}. CANARY is at or above MSP on almost every cell, the gains shrinking to seed-noise ties on ViT-L ($-0.0002$ AUROC on its Noise group).}
\label{tab:vitscale-cmp}
\resizebox{\textwidth}{!}{
\begin{tabular}{ll cc cc cc cc cc cc}
\toprule
 & & \multicolumn{8}{c}{\textbf{ImageNet-C}} & \multicolumn{2}{c}{\multirow{2}{*}{\textbf{ImageNet-A}}} & \multicolumn{2}{c}{\multirow{2}{*}{\textbf{ImageNet-R}}} \\
\cmidrule(lr){3-10}
 & & \multicolumn{2}{c}{\textbf{Noise}} & \multicolumn{2}{c}{\textbf{Blur}} & \multicolumn{2}{c}{\textbf{Weather}} & \multicolumn{2}{c}{\textbf{Digital}} & \multicolumn{2}{c}{} & \multicolumn{2}{c}{} \\
\cmidrule(lr){3-4}\cmidrule(lr){5-6}\cmidrule(lr){7-8}\cmidrule(lr){9-10}\cmidrule(lr){11-12}\cmidrule(lr){13-14}
\textbf{Backbone} & \textbf{Method} & e-AURC$\downarrow$ & AUROC$\uparrow$ & e-AURC$\downarrow$ & AUROC$\uparrow$ & e-AURC$\downarrow$ & AUROC$\uparrow$ & e-AURC$\downarrow$ & AUROC$\uparrow$ & e-AURC$\downarrow$ & AUROC$\uparrow$ & e-AURC$\downarrow$ & AUROC$\uparrow$ \\
\midrule
\multirow{2}{*}{ViT-S/16} & MSP & 67.6 & 0.8529 & 81.0 & 0.8426 & 65.0 & 0.8512 & 62.7 & 0.8492 & 209.7 & 0.6556 & 126.7 & 0.8196 \\
 & CANARY & \textbf{67.3} & \textbf{0.8533} & \textbf{80.4} & \textbf{0.8434} & \textbf{64.2} & \textbf{0.8524} & \textbf{61.9} & \textbf{0.8505} & \textbf{208.7} & \textbf{0.6577} & \textbf{122.7} & \textbf{0.8243} \\
\midrule
\multirow{2}{*}{\textbf{ViT-B/16}} & MSP & 53.9 & 0.8575 & 64.4 & 0.8531 & 50.5 & 0.8598 & 51.2 & 0.8551 & 235.7 & 0.6842 & 114.6 & 0.8272 \\
 & CANARY & \textbf{53.8} & \textbf{0.8577} & \textbf{64.1} & \textbf{0.8536} & \textbf{50.1} & \textbf{0.8606} & \textbf{50.5} & \textbf{0.8565} & \textbf{234.5} & \textbf{0.6865} & \textbf{113.5} & \textbf{0.8289} \\
\midrule
\multirow{2}{*}{ViT-L/16} & MSP & \textbf{41.7} & \textbf{0.8663} & 51.6 & 0.8617 & 43.1 & 0.8611 & 48.0 & 0.8463 & 206.6 & 0.7236 & 105.1 & 0.8325 \\
 & CANARY & 41.9 & 0.8661 & \textbf{51.3} & \textbf{0.8624} & \textbf{42.9} & \textbf{0.8619} & \textbf{47.2} & \textbf{0.8486} & \textbf{204.5} & \textbf{0.7262} & \textbf{103.6} & \textbf{0.8347} \\
\bottomrule
\end{tabular}
}
\end{table}

\begin{table}[htbp]
\centering
\caption{\textbf{ResNet-50.}}
\label{tab:full-resnet50}
\resizebox{\textwidth}{!}{
\begin{tabular}{l cc cc cc cc cc cc}
\toprule
\multirow{3}{*}{\textbf{Method}} & \multicolumn{8}{c}{\textbf{ImageNet-C}} & \multicolumn{2}{c}{\multirow{2}{*}{\textbf{ImageNet-A}}} & \multicolumn{2}{c}{\multirow{2}{*}{\textbf{ImageNet-R}}} \\
\cmidrule(lr){2-9}
 & \multicolumn{2}{c}{\textbf{Noise}} & \multicolumn{2}{c}{\textbf{Blur}} & \multicolumn{2}{c}{\textbf{Weather}} & \multicolumn{2}{c}{\textbf{Digital}} & \multicolumn{2}{c}{} & \multicolumn{2}{c}{} \\
\cmidrule(lr){2-3}\cmidrule(lr){4-5}\cmidrule(lr){6-7}\cmidrule(lr){8-9}\cmidrule(lr){10-11}\cmidrule(lr){12-13}
 & \textbf{e-AURC}$\downarrow$ & \textbf{AUROC}$\uparrow$ & \textbf{e-AURC}$\downarrow$ & \textbf{AUROC}$\uparrow$ & \textbf{e-AURC}$\downarrow$ & \textbf{AUROC}$\uparrow$ & \textbf{e-AURC}$\downarrow$ & \textbf{AUROC}$\uparrow$ & \textbf{e-AURC}$\downarrow$ & \textbf{AUROC}$\uparrow$ & \textbf{e-AURC}$\downarrow$ & \textbf{AUROC}$\uparrow$ \\
\midrule
\multicolumn{13}{l}{\textit{Logit-based}} \\
\midrule
MSP & \underline{79.0} & \underline{0.8515} & \underline{95.3} & \underline{0.8281} & 73.9 & 0.8493 & 75.4 & 0.8448 & 129.8 & 0.5538 & 126.9 & 0.8244 \\
MaxLogit & 83.6 & 0.8431 & 101.8 & 0.8169 & 81.7 & 0.8362 & 80.6 & 0.8339 & 130.3 & 0.5404 & 137.3 & 0.8094 \\
Energy & 93.7 & 0.8133 & 113.0 & 0.7826 & 88.1 & 0.8183 & 87.7 & 0.8128 & 131.8 & 0.5165 & 143.4 & 0.7912 \\
Doctor & \underline{78.8} & \underline{0.8518} & 95.4 & 0.8277 & \underline{73.7} & \underline{0.8498} & \underline{75.2} & \underline{0.8450} & 129.5 & \underline{0.5583} & 126.2 & 0.8262 \\
GEN & 83.6 & 0.8418 & 100.1 & 0.8181 & 77.1 & 0.8416 & 80.1 & 0.8342 & \underline{128.1} & \textbf{0.5677} & 124.6 & \underline{0.8290} \\
\midrule
\multicolumn{13}{l}{\textit{Activation-shaping}} \\
\midrule
ASH & 132.8 & 0.7611 & 155.7 & 0.7304 & 140.5 & 0.7371 & 132.9 & 0.7403 & \textbf{126.4} & \underline{0.5628} & 208.9 & 0.7177 \\
ReAct & 490.3 & 0.2407 & 466.2 & 0.2497 & 487.4 & 0.2380 & 468.6 & 0.2512 & 147.2 & 0.4089 & 501.8 & 0.2325 \\
DICE & 96.0 & 0.8079 & 116.2 & 0.7750 & 90.2 & 0.8135 & 89.7 & 0.8078 & 131.3 & 0.5223 & 143.5 & 0.7907 \\
\midrule
\multicolumn{13}{l}{\textit{Feature-based}} \\
\midrule
ViM & 225.2 & 0.6230 & 220.3 & 0.6423 & 195.6 & 0.6539 & 204.8 & 0.6380 & 134.4 & 0.5047 & 194.1 & 0.7153 \\
Deep kNN & 212.4 & 0.6517 & 232.2 & 0.6230 & 221.9 & 0.6208 & 213.9 & 0.6261 & 140.1 & 0.4637 & 260.6 & 0.6247 \\
MDS & 146.7 & 0.7401 & 156.7 & 0.7296 & 130.4 & 0.7507 & 138.7 & 0.7361 & 146.3 & 0.4112 & 151.3 & 0.7542 \\
MDS++ & 125.5 & 0.7718 & 142.6 & 0.7437 & 119.9 & 0.7621 & 121.3 & 0.7580 & 149.4 & 0.3769 & 173.2 & 0.7191 \\
RMDS & 127.2 & 0.7629 & 143.3 & 0.7386 & 113.9 & 0.7684 & 116.8 & 0.7600 & 148.0 & 0.3876 & 160.5 & 0.7462 \\
RMDS++ & 117.8 & 0.7799 & 135.1 & 0.7510 & 110.8 & 0.7755 & 108.3 & 0.7750 & 149.6 & 0.3648 & 182.3 & 0.7110 \\
NNGuide & 93.6 & 0.8155 & 113.4 & 0.7821 & 88.5 & 0.8178 & 88.1 & 0.8126 & 131.7 & 0.5161 & 149.5 & 0.7877 \\
fDBD & 89.2 & 0.8322 & 100.1 & 0.8209 & 79.5 & 0.8355 & 83.3 & 0.8283 & \underline{129.3} & 0.5518 & \textbf{107.7} & \textbf{0.8374} \\
TrustScore & 106.8 & 0.7978 & 126.2 & 0.7678 & 94.8 & 0.8038 & 101.5 & 0.7882 & 142.8 & 0.4204 & 150.1 & 0.7803 \\
\midrule
\multicolumn{13}{l}{\textit{Softmax-combination}} \\
\midrule
SIRC (L1) & 79.1 & 0.8512 & 96.1 & 0.8254 & 74.2 & 0.8488 & 75.6 & 0.8440 & 129.8 & 0.5546 & 127.1 & 0.8243 \\
SIRC (Residual) & 79.2 & 0.8499 & \textbf{93.6} & \textbf{0.8315} & \textbf{72.5} & \textbf{0.8520} & \textbf{74.5} & \textbf{0.8466} & 129.8 & 0.5537 & \underline{123.7} & 0.8289 \\
\midrule
\rowcolor{gray!15} \textbf{CANARY (ours)} & \textbf{78.6} & \textbf{0.8518} & \underline{94.2} & \underline{0.8296} & \underline{72.7} & \underline{0.8508} & \underline{74.6} & \underline{0.8458} & 129.4 & 0.5562 & \underline{121.4} & \underline{0.8306} \\
{\scriptsize ($\Delta$ vs.\ MSP)} & {\scriptsize 0.4$\uparrow$} & {\scriptsize 0.0003$\uparrow$} & {\scriptsize 1.1$\uparrow$} & {\scriptsize 0.0015$\uparrow$} & {\scriptsize 1.2$\uparrow$} & {\scriptsize 0.0015$\uparrow$} & {\scriptsize 0.7$\uparrow$} & {\scriptsize 0.0011$\uparrow$} & {\scriptsize 0.4$\uparrow$} & {\scriptsize 0.0025$\uparrow$} & {\scriptsize 5.5$\uparrow$} & {\scriptsize 0.0062$\uparrow$} \\
\bottomrule
\end{tabular}
}
\end{table}

\begin{table}[htbp]
\centering
\caption{\textbf{Swin-B.}}
\label{tab:full-swinb}
\resizebox{\textwidth}{!}{
\begin{tabular}{l cc cc cc cc cc cc}
\toprule
\multirow{3}{*}{\textbf{Method}} & \multicolumn{8}{c}{\textbf{ImageNet-C}} & \multicolumn{2}{c}{\multirow{2}{*}{\textbf{ImageNet-A}}} & \multicolumn{2}{c}{\multirow{2}{*}{\textbf{ImageNet-R}}} \\
\cmidrule(lr){2-9}
 & \multicolumn{2}{c}{\textbf{Noise}} & \multicolumn{2}{c}{\textbf{Blur}} & \multicolumn{2}{c}{\textbf{Weather}} & \multicolumn{2}{c}{\textbf{Digital}} & \multicolumn{2}{c}{} & \multicolumn{2}{c}{} \\
\cmidrule(lr){2-3}\cmidrule(lr){4-5}\cmidrule(lr){6-7}\cmidrule(lr){8-9}\cmidrule(lr){10-11}\cmidrule(lr){12-13}
 & \textbf{e-AURC}$\downarrow$ & \textbf{AUROC}$\uparrow$ & \textbf{e-AURC}$\downarrow$ & \textbf{AUROC}$\uparrow$ & \textbf{e-AURC}$\downarrow$ & \textbf{AUROC}$\uparrow$ & \textbf{e-AURC}$\downarrow$ & \textbf{AUROC}$\uparrow$ & \textbf{e-AURC}$\downarrow$ & \textbf{AUROC}$\uparrow$ & \textbf{e-AURC}$\downarrow$ & \textbf{AUROC}$\uparrow$ \\
\midrule
\multicolumn{13}{l}{\textit{Logit-based}} \\
\midrule
MSP & 65.9 & \underline{0.8438} & 80.3 & \underline{0.8391} & 70.1 & \underline{0.8197} & 66.9 & 0.8286 & 250.2 & 0.6685 & 128.3 & 0.8197 \\
MaxLogit & 109.2 & 0.7802 & 121.5 & 0.7880 & 106.0 & 0.7585 & 109.9 & 0.7563 & 292.5 & 0.6152 & 179.2 & 0.7744 \\
Energy & 163.1 & 0.6710 & 171.9 & 0.6934 & 159.2 & 0.6400 & 157.7 & 0.6481 & 332.4 & 0.5388 & 223.3 & 0.7059 \\
Doctor & 66.7 & 0.8413 & 81.0 & 0.8370 & 73.6 & 0.8099 & 68.7 & 0.8226 & 252.3 & 0.6639 & 128.2 & 0.8198 \\
GEN & 86.6 & 0.8002 & 95.2 & 0.8085 & 98.4 & 0.7553 & 89.1 & 0.7740 & 274.5 & 0.6393 & 124.6 & 0.8194 \\
\midrule
\multicolumn{13}{l}{\textit{Activation-shaping}} \\
\midrule
ASH & 162.5 & 0.6784 & 176.5 & 0.6902 & 158.1 & 0.6454 & 158.6 & 0.6555 & 324.8 & 0.5592 & 265.6 & 0.6609 \\
ReAct & 116.1 & 0.7355 & 129.0 & 0.7393 & 123.4 & 0.6951 & 110.2 & 0.7217 & 317.5 & 0.5578 & 150.9 & 0.7666 \\
DICE & 206.4 & 0.5811 & 216.6 & 0.6059 & 195.1 & 0.5634 & 191.0 & 0.5754 & 362.7 & 0.4861 & 281.6 & 0.6187 \\
\midrule
\multicolumn{13}{l}{\textit{Feature-based}} \\
\midrule
ViM & 106.4 & 0.7577 & 111.6 & 0.7745 & 105.6 & 0.7369 & 96.4 & 0.7575 & 284.9 & 0.6217 & 127.4 & 0.7930 \\
Deep kNN & 141.0 & 0.7244 & 136.5 & 0.7581 & 133.3 & 0.7067 & 133.3 & 0.7122 & 281.6 & 0.6333 & 119.0 & 0.8246 \\
MDS & 88.5 & 0.7850 & 97.0 & 0.7950 & 88.2 & 0.7699 & 79.3 & 0.7853 & 270.5 & 0.6406 & 111.9 & 0.8137 \\
MDS++ & 87.5 & 0.7906 & 98.3 & 0.7952 & 92.5 & 0.7611 & 80.0 & 0.7861 & 326.3 & 0.5434 & 127.3 & 0.7797 \\
RMDS & 78.0 & 0.8082 & 86.9 & 0.8146 & 75.1 & 0.8000 & 69.2 & 0.8101 & 258.6 & 0.6598 & \underline{101.1} & \underline{0.8306} \\
RMDS++ & 81.4 & 0.8009 & 93.6 & 0.8002 & 84.4 & 0.7767 & 73.8 & 0.7982 & 334.4 & 0.5290 & 134.6 & 0.7637 \\
NNGuide & 87.0 & 0.7911 & 95.0 & 0.8008 & 90.2 & 0.7669 & 82.6 & 0.7802 & 320.5 & 0.5717 & 128.4 & 0.7852 \\
fDBD & 78.1 & 0.8079 & 86.0 & 0.8170 & 78.9 & 0.7926 & 71.9 & 0.8032 & \underline{245.0} & \underline{0.6771} & \textbf{91.7} & \textbf{0.8534} \\
TrustScore & \underline{65.0} & 0.8282 & \underline{77.0} & 0.8260 & \textbf{58.0} & \textbf{0.8305} & \textbf{57.3} & \underline{0.8315} & \underline{246.4} & \underline{0.6770} & 122.1 & 0.7992 \\
\midrule
\multicolumn{13}{l}{\textit{Softmax-combination}} \\
\midrule
SIRC (L1) & 67.6 & 0.8415 & 81.8 & 0.8374 & 71.6 & 0.8170 & 69.0 & 0.8246 & 251.1 & 0.6667 & 130.9 & 0.8174 \\
SIRC (Residual) & \underline{62.7} & \underline{0.8434} & \underline{75.1} & \underline{0.8398} & \underline{68.6} & 0.8173 & \underline{59.8} & \underline{0.8371} & 248.0 & 0.6668 & 110.0 & 0.8297 \\
\midrule
\rowcolor{gray!15} \textbf{CANARY (ours)} & \textbf{59.3} & \textbf{0.8507} & \textbf{71.8} & \textbf{0.8460} & \underline{64.8} & \underline{0.8262} & \underline{58.1} & \textbf{0.8391} & \textbf{244.0} & \textbf{0.6798} & \underline{102.4} & \underline{0.8388} \\
{\scriptsize ($\Delta$ vs.\ MSP)} & {\scriptsize 6.7$\uparrow$} & {\scriptsize 0.0068$\uparrow$} & {\scriptsize 8.5$\uparrow$} & {\scriptsize 0.0069$\uparrow$} & {\scriptsize 5.2$\uparrow$} & {\scriptsize 0.0065$\uparrow$} & {\scriptsize 8.8$\uparrow$} & {\scriptsize 0.0105$\uparrow$} & {\scriptsize 6.2$\uparrow$} & {\scriptsize 0.0113$\uparrow$} & {\scriptsize 25.8$\uparrow$} & {\scriptsize 0.0191$\uparrow$} \\
\bottomrule
\end{tabular}
}
\end{table}

\begin{table}[htbp]
\centering
\caption{\textbf{ConvNeXt-B.}}
\label{tab:full-convnextb}
\resizebox{\textwidth}{!}{
\begin{tabular}{l cc cc cc cc cc cc}
\toprule
\multirow{3}{*}{\textbf{Method}} & \multicolumn{8}{c}{\textbf{ImageNet-C}} & \multicolumn{2}{c}{\multirow{2}{*}{\textbf{ImageNet-A}}} & \multicolumn{2}{c}{\multirow{2}{*}{\textbf{ImageNet-R}}} \\
\cmidrule(lr){2-9}
 & \multicolumn{2}{c}{\textbf{Noise}} & \multicolumn{2}{c}{\textbf{Blur}} & \multicolumn{2}{c}{\textbf{Weather}} & \multicolumn{2}{c}{\textbf{Digital}} & \multicolumn{2}{c}{} & \multicolumn{2}{c}{} \\
\cmidrule(lr){2-3}\cmidrule(lr){4-5}\cmidrule(lr){6-7}\cmidrule(lr){8-9}\cmidrule(lr){10-11}\cmidrule(lr){12-13}
 & \textbf{e-AURC}$\downarrow$ & \textbf{AUROC}$\uparrow$ & \textbf{e-AURC}$\downarrow$ & \textbf{AUROC}$\uparrow$ & \textbf{e-AURC}$\downarrow$ & \textbf{AUROC}$\uparrow$ & \textbf{e-AURC}$\downarrow$ & \textbf{AUROC}$\uparrow$ & \textbf{e-AURC}$\downarrow$ & \textbf{AUROC}$\uparrow$ & \textbf{e-AURC}$\downarrow$ & \textbf{AUROC}$\uparrow$ \\
\midrule
\multicolumn{13}{l}{\textit{Logit-based}} \\
\midrule
MSP & \underline{55.5} & \underline{0.8572} & \underline{72.1} & \underline{0.8490} & \underline{50.7} & \underline{0.8630} & 55.2 & \underline{0.8527} & 226.6 & 0.6982 & 109.3 & 0.8373 \\
MaxLogit & 85.3 & 0.8099 & 99.3 & 0.8132 & 77.9 & 0.8170 & 84.3 & 0.8003 & 263.4 & 0.6616 & 129.8 & 0.8209 \\
Energy & 110.0 & 0.7578 & 122.6 & 0.7700 & 98.7 & 0.7698 & 107.4 & 0.7469 & 287.3 & 0.6207 & 143.6 & 0.8005 \\
Doctor & 55.8 & \underline{0.8561} & 72.5 & \underline{0.8477} & 50.9 & 0.8622 & 55.5 & 0.8515 & 227.3 & 0.6964 & 108.8 & 0.8381 \\
GEN & 75.3 & 0.8161 & 87.7 & 0.8186 & 67.7 & 0.8241 & 74.6 & 0.8072 & 251.9 & 0.6699 & 106.6 & 0.8375 \\
\midrule
\multicolumn{13}{l}{\textit{Activation-shaping}} \\
\midrule
ASH & 104.9 & 0.7595 & 116.3 & 0.7728 & 95.5 & 0.7666 & 105.6 & 0.7436 & 279.0 & 0.6344 & 157.4 & 0.7795 \\
ReAct & 88.5 & 0.7900 & 102.0 & 0.7938 & 78.7 & 0.8010 & 86.1 & 0.7820 & 273.5 & 0.6296 & 117.3 & 0.8198 \\
DICE & 134.4 & 0.7030 & 146.6 & 0.7241 & 121.5 & 0.7175 & 129.3 & 0.6945 & 325.7 & 0.5671 & 181.4 & 0.7525 \\
\midrule
\multicolumn{13}{l}{\textit{Feature-based}} \\
\midrule
ViM & 93.2 & 0.7717 & 104.6 & 0.7818 & 81.6 & 0.7844 & 86.8 & 0.7708 & 266.4 & 0.6386 & 114.2 & 0.8098 \\
Deep kNN & 146.6 & 0.7196 & 145.9 & 0.7516 & 134.4 & 0.7250 & 145.7 & 0.7024 & 263.9 & 0.6667 & 115.9 & 0.8420 \\
MDS & 75.6 & 0.8046 & 88.2 & 0.8093 & 66.9 & 0.8148 & 70.0 & 0.8043 & 246.6 & 0.6644 & 94.5 & 0.8376 \\
MDS++ & 77.0 & 0.8024 & 90.9 & 0.8045 & 69.1 & 0.8097 & 72.0 & 0.8005 & 278.0 & 0.6181 & 106.7 & 0.8136 \\
RMDS & 68.0 & 0.8203 & 80.8 & 0.8221 & 60.2 & 0.8295 & 62.7 & 0.8215 & 240.5 & 0.6742 & \underline{91.9} & 0.8406 \\
RMDS++ & 70.7 & 0.8138 & 85.5 & 0.8122 & 63.7 & 0.8199 & 65.7 & 0.8134 & 282.4 & 0.6116 & 110.5 & 0.8041 \\
NNGuide & 74.9 & 0.8094 & 88.0 & 0.8114 & 66.5 & 0.8184 & 72.6 & 0.8032 & 274.2 & 0.6376 & 108.5 & 0.8200 \\
fDBD & 65.1 & 0.8287 & 77.2 & 0.8308 & 58.0 & 0.8372 & 61.4 & 0.8265 & \underline{211.4} & \underline{0.7160} & \textbf{78.7} & \textbf{0.8677} \\
TrustScore & 58.3 & 0.8373 & 75.5 & 0.8272 & 52.0 & 0.8458 & \underline{55.0} & 0.8385 & \textbf{200.3} & \textbf{0.7344} & 102.6 & 0.8294 \\
\midrule
\multicolumn{13}{l}{\textit{Softmax-combination}} \\
\midrule
SIRC (L1) & 56.8 & 0.8555 & 73.4 & 0.8477 & 51.9 & 0.8614 & 56.5 & 0.8509 & 227.0 & 0.6977 & 111.1 & 0.8361 \\
SIRC (Residual) & \underline{54.4} & 0.8560 & \underline{70.0} & 0.8477 & \underline{48.6} & \underline{0.8641} & \underline{52.5} & \underline{0.8544} & 225.7 & 0.6946 & 100.2 & \underline{0.8431} \\
\midrule
\rowcolor{gray!15} \textbf{CANARY (ours)} & \textbf{52.2} & \textbf{0.8603} & \textbf{67.2} & \textbf{0.8529} & \textbf{46.6} & \textbf{0.8675} & \textbf{50.6} & \textbf{0.8579} & \underline{215.0} & \underline{0.7117} & \underline{91.8} & \underline{0.8519} \\
{\scriptsize ($\Delta$ vs.\ MSP)} & {\scriptsize 3.3$\uparrow$} & {\scriptsize 0.0031$\uparrow$} & {\scriptsize 4.8$\uparrow$} & {\scriptsize 0.0039$\uparrow$} & {\scriptsize 4.0$\uparrow$} & {\scriptsize 0.0044$\uparrow$} & {\scriptsize 4.5$\uparrow$} & {\scriptsize 0.0051$\uparrow$} & {\scriptsize 11.6$\uparrow$} & {\scriptsize 0.0135$\uparrow$} & {\scriptsize 17.5$\uparrow$} & {\scriptsize 0.0146$\uparrow$} \\
\bottomrule
\end{tabular}
}
\end{table}

\begin{table}[htbp]
\centering
\caption{\textbf{EfficientNet-B4.}}
\label{tab:full-efficientnetb4}
\resizebox{\textwidth}{!}{
\begin{tabular}{l cc cc cc cc cc cc}
\toprule
\multirow{3}{*}{\textbf{Method}} & \multicolumn{8}{c}{\textbf{ImageNet-C}} & \multicolumn{2}{c}{\multirow{2}{*}{\textbf{ImageNet-A}}} & \multicolumn{2}{c}{\multirow{2}{*}{\textbf{ImageNet-R}}} \\
\cmidrule(lr){2-9}
 & \multicolumn{2}{c}{\textbf{Noise}} & \multicolumn{2}{c}{\textbf{Blur}} & \multicolumn{2}{c}{\textbf{Weather}} & \multicolumn{2}{c}{\textbf{Digital}} & \multicolumn{2}{c}{} & \multicolumn{2}{c}{} \\
\cmidrule(lr){2-3}\cmidrule(lr){4-5}\cmidrule(lr){6-7}\cmidrule(lr){8-9}\cmidrule(lr){10-11}\cmidrule(lr){12-13}
 & \textbf{e-AURC}$\downarrow$ & \textbf{AUROC}$\uparrow$ & \textbf{e-AURC}$\downarrow$ & \textbf{AUROC}$\uparrow$ & \textbf{e-AURC}$\downarrow$ & \textbf{AUROC}$\uparrow$ & \textbf{e-AURC}$\downarrow$ & \textbf{AUROC}$\uparrow$ & \textbf{e-AURC}$\downarrow$ & \textbf{AUROC}$\uparrow$ & \textbf{e-AURC}$\downarrow$ & \textbf{AUROC}$\uparrow$ \\
\midrule
\multicolumn{13}{l}{\textit{Logit-based}} \\
\midrule
MSP & \underline{79.8} & \underline{0.8473} & \underline{97.5} & \underline{0.8261} & 78.2 & \underline{0.8388} & 82.8 & \underline{0.8303} & 230.6 & \underline{0.6240} & 140.4 & 0.8133 \\
MaxLogit & 109.9 & 0.8097 & 127.1 & 0.7931 & 120.0 & 0.7813 & 122.4 & 0.7739 & 241.2 & 0.6005 & 192.9 & 0.7655 \\
Energy & 164.5 & 0.7126 & 181.9 & 0.6989 & 183.7 & 0.6577 & 187.0 & 0.6466 & 259.1 & 0.5418 & 265.0 & 0.6513 \\
Doctor & 80.7 & 0.8451 & 98.8 & 0.8230 & 79.1 & 0.8362 & 85.0 & 0.8251 & 230.8 & 0.6232 & 140.1 & 0.8139 \\
GEN & 96.7 & 0.8184 & 112.1 & 0.8028 & 93.1 & 0.8086 & 103.1 & 0.7928 & 234.4 & 0.6205 & 124.8 & 0.8273 \\
\midrule
\multicolumn{13}{l}{\textit{Activation-shaping}} \\
\midrule
ASH & 223.7 & 0.6118 & 238.5 & 0.6016 & 243.8 & 0.5471 & 225.5 & 0.5682 & 292.1 & 0.4613 & 365.7 & 0.5045 \\
ReAct & 177.4 & 0.6703 & 183.2 & 0.6789 & 159.1 & 0.6667 & 176.5 & 0.6437 & 249.2 & 0.5698 & 195.2 & 0.6938 \\
DICE & 235.2 & 0.5854 & 245.3 & 0.5874 & 261.9 & 0.5103 & 256.6 & 0.5135 & 271.0 & 0.5037 & 362.7 & 0.4957 \\
\midrule
\multicolumn{13}{l}{\textit{Feature-based}} \\
\midrule
ViM & 129.8 & 0.7467 & 138.5 & 0.7410 & 101.4 & 0.7789 & 118.8 & 0.7530 & 247.3 & 0.5837 & 125.5 & 0.8111 \\
Deep kNN & 133.3 & 0.7683 & 138.3 & 0.7657 & 125.3 & 0.7640 & 136.7 & 0.7470 & 237.6 & 0.6051 & \underline{115.5} & 0.8292 \\
MDS & 114.2 & 0.7720 & 124.7 & 0.7632 & 90.5 & 0.7965 & 102.8 & 0.7778 & 252.2 & 0.5731 & 118.3 & 0.8184 \\
MDS++ & 104.2 & 0.7931 & 118.3 & 0.7782 & 86.9 & 0.8021 & 96.2 & 0.7880 & 281.5 & 0.4723 & 137.9 & 0.7664 \\
RMDS & 97.7 & 0.8041 & 110.3 & 0.7903 & 79.1 & 0.8190 & 87.4 & 0.8069 & 257.0 & 0.5602 & 117.3 & 0.8177 \\
RMDS++ & 105.6 & 0.7888 & 120.6 & 0.7712 & 86.3 & 0.8015 & 94.4 & 0.7895 & 294.3 & 0.4434 & 151.6 & 0.7425 \\
NNGuide & 104.9 & 0.7946 & 123.5 & 0.7681 & 92.3 & 0.7974 & 104.0 & 0.7795 & 273.4 & 0.5187 & 139.8 & 0.7794 \\
fDBD & 93.4 & 0.8194 & 104.9 & 0.8081 & 83.2 & 0.8178 & 91.4 & 0.8059 & \underline{230.5} & \underline{0.6371} & \textbf{100.4} & \textbf{0.8510} \\
TrustScore & 90.4 & 0.8105 & 106.0 & 0.7925 & \underline{73.6} & 0.8227 & \underline{79.6} & 0.8137 & 238.8 & 0.6148 & 123.4 & 0.8091 \\
\midrule
\multicolumn{13}{l}{\textit{Softmax-combination}} \\
\midrule
SIRC (L1) & 81.2 & \underline{0.8456} & 99.2 & \underline{0.8239} & 80.5 & 0.8355 & 86.1 & 0.8244 & \underline{230.5} & 0.6235 & 144.7 & 0.8090 \\
SIRC (Residual) & \underline{79.7} & 0.8398 & \underline{95.4} & 0.8184 & \underline{70.2} & \underline{0.8464} & \underline{77.1} & \underline{0.8346} & 230.7 & 0.6205 & 118.9 & \underline{0.8344} \\
\midrule
\rowcolor{gray!15} \textbf{CANARY (ours)} & \textbf{73.8} & \textbf{0.8518} & \textbf{89.4} & \textbf{0.8323} & \textbf{65.7} & \textbf{0.8512} & \textbf{73.3} & \textbf{0.8398} & \textbf{225.1} & \textbf{0.6396} & \underline{108.9} & \underline{0.8413} \\
{\scriptsize ($\Delta$ vs.\ MSP)} & {\scriptsize 6.0$\uparrow$} & {\scriptsize 0.0045$\uparrow$} & {\scriptsize 8.1$\uparrow$} & {\scriptsize 0.0062$\uparrow$} & {\scriptsize 12.5$\uparrow$} & {\scriptsize 0.0124$\uparrow$} & {\scriptsize 9.5$\uparrow$} & {\scriptsize 0.0095$\uparrow$} & {\scriptsize 5.5$\uparrow$} & {\scriptsize 0.0156$\uparrow$} & {\scriptsize 31.5$\uparrow$} & {\scriptsize 0.0280$\uparrow$} \\
\bottomrule
\end{tabular}
}
\end{table}

\subsection{ObjectNet}
\label{app:objectnet}

Table~\ref{tab:objectnet} reports every method on the held-out ObjectNet set (16{,}825 images). CANARY matches or exceeds MSP on ObjectNet across all five backbones, and no baseline shares this never-worse property.

\begin{table}[htbp]
\centering
\caption{\textbf{ObjectNet.} Failure prediction on ObjectNet across the five main backbones. e-AURC $\times 10^3$; best per column in \textbf{bold}, 2nd/3rd \underline{underlined}. The bottom row ($\Delta$ vs.\ MSP) is CANARY${-}$MSP ($\uparrow$ = CANARY improves over MSP, $\downarrow$ = worse).}
\label{tab:objectnet}
\resizebox{\textwidth}{!}{
\begin{tabular}{l cc cc cc cc cc}
\toprule
\multirow{2}{*}{\textbf{Method}} & \multicolumn{2}{c}{\textbf{ViT-B/16}} & \multicolumn{2}{c}{\textbf{ResNet-50}} & \multicolumn{2}{c}{\textbf{Swin-B}} & \multicolumn{2}{c}{\textbf{ConvNeXt-B}} & \multicolumn{2}{c}{\textbf{EfficientNet-B4}} \\
\cmidrule(lr){2-3} \cmidrule(lr){4-5} \cmidrule(lr){6-7} \cmidrule(lr){8-9} \cmidrule(lr){10-11}
 & \textbf{e-AURC}$\downarrow$ & \textbf{AUROC}$\uparrow$ & \textbf{e-AURC}$\downarrow$ & \textbf{AUROC}$\uparrow$ & \textbf{e-AURC}$\downarrow$ & \textbf{AUROC}$\uparrow$ & \textbf{e-AURC}$\downarrow$ & \textbf{AUROC}$\uparrow$ & \textbf{e-AURC}$\downarrow$ & \textbf{AUROC}$\uparrow$ \\
\midrule
\multicolumn{11}{l}{\textit{Logit-based}} \\
\midrule
MSP & \underline{103.2} & \underline{0.8217} & 120.1 & 0.8104 & 134.1 & 0.7941 & 106.5 & 0.8181 & 129.3 & 0.8023 \\
MaxLogit & 120.3 & 0.7970 & 131.3 & 0.7947 & 197.8 & 0.7343 & 141.9 & 0.7859 & 176.4 & 0.7540 \\
Energy & 134.1 & 0.7717 & 139.6 & 0.7723 & 256.8 & 0.6434 & 164.7 & 0.7503 & 240.6 & 0.6474 \\
Doctor & \textbf{102.8} & \textbf{0.8220} & 119.5 & 0.8118 & 134.7 & 0.7927 & 106.0 & 0.8189 & 130.0 & 0.8011 \\
GEN & 116.8 & 0.8017 & 122.1 & 0.8078 & 144.6 & 0.7791 & 113.0 & 0.8081 & 129.2 & 0.8015 \\
\midrule
\multicolumn{11}{l}{\textit{Activation-shaping}} \\
\midrule
ASH & 224.9 & 0.6500 & 256.1 & 0.6415 & 295.8 & 0.5975 & 165.7 & 0.7406 & 323.5 & 0.5454 \\
ReAct & 133.8 & 0.7728 & 505.6 & 0.2601 & 177.0 & 0.7117 & 138.7 & 0.7687 & 306.3 & 0.5428 \\
DICE & 144.5 & 0.7561 & 141.2 & 0.7687 & 295.7 & 0.5752 & 183.3 & 0.7190 & 300.3 & 0.5505 \\
\midrule
\multicolumn{11}{l}{\textit{Feature-based}} \\
\midrule
ViM & 261.5 & 0.6087 & 307.0 & 0.5709 & 133.9 & 0.7767 & 135.8 & 0.7593 & 154.7 & 0.7611 \\
Deep kNN & 371.4 & 0.4998 & 466.0 & 0.3491 & 251.2 & 0.6743 & 300.8 & 0.5760 & 219.4 & 0.7089 \\
MDS & 122.5 & 0.8000 & 162.8 & 0.7533 & 114.1 & 0.8020 & 99.0 & 0.8171 & 132.4 & 0.7853 \\
MDS++ & 130.6 & 0.7871 & 159.2 & 0.7445 & 115.8 & 0.8019 & 99.0 & 0.8179 & 123.2 & 0.7939 \\
RMDS & 108.0 & 0.8193 & 133.5 & 0.7821 & \textbf{101.1} & \textbf{0.8220} & \textbf{89.8} & \textbf{0.8306} & 114.5 & 0.8080 \\
RMDS++ & 108.3 & 0.8180 & 125.0 & 0.7943 & \underline{105.4} & \underline{0.8142} & \underline{90.8} & \underline{0.8285} & 117.8 & 0.7977 \\
NNGuide & 137.1 & 0.7691 & 136.5 & 0.7748 & 133.8 & 0.7774 & 106.6 & 0.8074 & 132.1 & 0.7857 \\
fDBD & 108.0 & 0.8166 & \textbf{117.7} & \textbf{0.8131} & \underline{104.5} & \underline{0.8181} & \underline{94.7} & 0.8250 & \underline{113.5} & \underline{0.8186} \\
TrustScore & 126.7 & 0.7791 & 141.7 & 0.7674 & 112.6 & 0.7956 & 109.3 & 0.7912 & 133.3 & 0.7711 \\
\midrule
\multicolumn{11}{l}{\textit{Softmax-combination}} \\
\midrule
SIRC (L1) & 105.0 & 0.8166 & 120.5 & 0.8095 & 136.8 & 0.7915 & 108.4 & 0.8166 & 131.9 & 0.7996 \\
SIRC (Residual) & 105.1 & 0.8166 & \underline{119.0} & \underline{0.8123} & 113.8 & 0.8128 & 100.8 & 0.8213 & \underline{114.1} & \underline{0.8176} \\
\midrule
\rowcolor{gray!15} \textbf{CANARY (ours)} & \underline{103.2} & \underline{0.8217} & \underline{118.4} & \underline{0.8121} & 113.3 & 0.8086 & 95.8 & \underline{0.8258} & \textbf{110.1} & \textbf{0.8197} \\
{\scriptsize ($\Delta$ vs.\ MSP)} & {\scriptsize 0.1$\uparrow$} & {\scriptsize 0.0000} & {\scriptsize 1.7$\uparrow$} & {\scriptsize 0.0018$\uparrow$} & {\scriptsize 20.8$\uparrow$} & {\scriptsize 0.0146$\uparrow$} & {\scriptsize 10.7$\uparrow$} & {\scriptsize 0.0077$\uparrow$} & {\scriptsize 19.2$\uparrow$} & {\scriptsize 0.0174$\uparrow$} \\
\bottomrule
\end{tabular}
}
\end{table}

\clearpage

\subsection{DINOv2 (self-supervised)}
\label{app:dinov2}

Table~\ref{tab:dinov2} reports all 20 detectors on DINOv2 on ImageNet-C by corruption group and ImageNet-A/R. e-AURC is reported $\times10^3$. Per column the best detector is in \textbf{bold} and the second/third \underline{underlined}. The bottom row ($\Delta$ vs.\ MSP) is CANARY${-}$MSP ($\uparrow$ = CANARY improves over MSP, $\downarrow$ = worse).

\begin{table}[htbp]
\centering
\caption{\textbf{DINOv2 (self-supervised).} Failure prediction on the self-supervised DINOv2 backbone: ImageNet-C by corruption group + ImageNet-A/R. Best per column \textbf{bold}, 2nd/3rd \underline{underlined}.}
\label{tab:dinov2}
\resizebox{\textwidth}{!}{
\begin{tabular}{l cc cc cc cc cc cc}
\toprule
\multirow{3}{*}{\textbf{Method}} & \multicolumn{8}{c}{\textbf{ImageNet-C}} & \multicolumn{2}{c}{\multirow{2}{*}{\textbf{ImageNet-A}}} & \multicolumn{2}{c}{\multirow{2}{*}{\textbf{ImageNet-R}}} \\
\cmidrule(lr){2-9}
 & \multicolumn{2}{c}{\textbf{Noise}} & \multicolumn{2}{c}{\textbf{Blur}} & \multicolumn{2}{c}{\textbf{Weather}} & \multicolumn{2}{c}{\textbf{Digital}} & \multicolumn{2}{c}{} & \multicolumn{2}{c}{} \\
\cmidrule(lr){2-3}\cmidrule(lr){4-5}\cmidrule(lr){6-7}\cmidrule(lr){8-9}\cmidrule(lr){10-11}\cmidrule(lr){12-13}
 & \textbf{e-AURC}$\downarrow$ & \textbf{AUROC}$\uparrow$ & \textbf{e-AURC}$\downarrow$ & \textbf{AUROC}$\uparrow$ & \textbf{e-AURC}$\downarrow$ & \textbf{AUROC}$\uparrow$ & \textbf{e-AURC}$\downarrow$ & \textbf{AUROC}$\uparrow$ & \textbf{e-AURC}$\downarrow$ & \textbf{AUROC}$\uparrow$ & \textbf{e-AURC}$\downarrow$ & \textbf{AUROC}$\uparrow$ \\
\midrule
\multicolumn{13}{l}{\textit{Logit-based}} \\
\midrule
MSP & 70.9 & 0.8426 & 70.4 & 0.8481 & 45.3 & 0.8632 & 47.9 & 0.8588 & 204.9 & 0.7290 & 104.4 & 0.8258 \\
MaxLogit & 93.0 & 0.7990 & 99.2 & 0.7999 & 75.5 & 0.7983 & 77.2 & 0.7899 & 237.0 & 0.6932 & 114.7 & 0.8132 \\
Energy & 100.4 & 0.7841 & 108.9 & 0.7812 & 83.9 & 0.7788 & 85.3 & 0.7698 & 249.5 & 0.6726 & 123.1 & 0.7994 \\
Doctor & 70.3 & 0.8437 & 70.5 & 0.8476 & 45.4 & 0.8625 & 48.0 & 0.8581 & 205.3 & 0.7278 & 104.0 & 0.8262 \\
GEN & 92.3 & 0.7976 & 98.3 & 0.7993 & 74.7 & 0.7975 & 76.5 & 0.7892 & 239.2 & 0.6900 & 112.6 & 0.8148 \\
\midrule
\multicolumn{13}{l}{\textit{Activation-shaping}} \\
\midrule
ASH & 144.0 & 0.7002 & 160.4 & 0.6913 & 128.0 & 0.6831 & 128.4 & 0.6731 & 302.4 & 0.5954 & 211.1 & 0.6793 \\
ReAct & 101.8 & 0.7815 & 111.2 & 0.7768 & 85.3 & 0.7752 & 86.8 & 0.7661 & 251.3 & 0.6675 & 125.2 & 0.7957 \\
DICE & 112.0 & 0.7610 & 122.4 & 0.7561 & 94.2 & 0.7545 & 96.2 & 0.7435 & 274.3 & 0.6378 & 156.9 & 0.7482 \\
\midrule
\multicolumn{13}{l}{\textit{Feature-based}} \\
\midrule
ViM & 143.8 & 0.6896 & 163.8 & 0.6754 & 124.9 & 0.6850 & 123.8 & 0.6796 & 313.8 & 0.5706 & 197.7 & 0.6658 \\
Deep kNN & 152.0 & 0.7090 & 165.5 & 0.7080 & 148.5 & 0.6783 & 149.3 & 0.6646 & 278.4 & 0.6291 & 137.2 & 0.7828 \\
MDS & 104.9 & 0.7646 & 110.4 & 0.7725 & 84.4 & 0.7771 & 84.9 & 0.7703 & 280.3 & 0.6126 & 127.0 & 0.7711 \\
MDS++ & 93.8 & 0.7924 & 100.2 & 0.7949 & 80.9 & 0.7884 & 81.2 & 0.7809 & 256.8 & 0.6471 & 104.2 & 0.8130 \\
RMDS & 99.0 & 0.7778 & 98.6 & 0.7940 & 75.6 & 0.7960 & 76.1 & 0.7898 & 276.9 & 0.6175 & 119.2 & 0.7834 \\
RMDS++ & 94.9 & 0.7886 & 94.3 & 0.8035 & 74.1 & 0.8006 & 74.5 & 0.7942 & 264.6 & 0.6384 & 108.5 & 0.8052 \\
NNGuide & 88.6 & 0.8030 & 96.2 & 0.8025 & 74.1 & 0.7995 & 76.3 & 0.7899 & 247.1 & 0.6775 & 110.0 & 0.8125 \\
fDBD & 93.6 & 0.7963 & 97.7 & 0.8003 & 77.7 & 0.7930 & 78.1 & 0.7864 & 211.4 & 0.7229 & \underline{99.6} & \underline{0.8399} \\
TrustScore & \underline{68.9} & \underline{0.8448} & \textbf{63.2} & \textbf{0.8595} & 47.2 & 0.8607 & 49.2 & 0.8559 & \textbf{199.8} & \underline{0.7330} & 105.7 & 0.8143 \\
\midrule
\multicolumn{13}{l}{\textit{Softmax-combination}} \\
\midrule
SIRC (L1) & \textbf{64.4} & \textbf{0.8530} & \underline{68.7} & \underline{0.8487} & \underline{44.5} & \underline{0.8641} & \textbf{46.6} & \underline{0.8603} & \underline{200.4} & \textbf{0.7345} & \textbf{93.7} & \textbf{0.8443} \\
SIRC (Residual) & \underline{67.7} & \underline{0.8475} & 70.7 & 0.8451 & \underline{44.7} & \underline{0.8641} & \underline{46.9} & \underline{0.8603} & 205.6 & 0.7251 & 102.9 & 0.8261 \\
\midrule
\rowcolor{gray!15} \textbf{CANARY (ours)} & 69.3 & 0.8447 & \underline{68.8} & \underline{0.8500} & \textbf{44.4} & \textbf{0.8650} & \underline{46.9} & \textbf{0.8607} & \underline{203.1} & \underline{0.7311} & \underline{101.2} & \underline{0.8289} \\
{\scriptsize ($\Delta$ vs.\ MSP)} & {\scriptsize 1.6$\uparrow$} & {\scriptsize 0.0021$\uparrow$} & {\scriptsize 1.6$\uparrow$} & {\scriptsize 0.0019$\uparrow$} & {\scriptsize 0.9$\uparrow$} & {\scriptsize 0.0018$\uparrow$} & {\scriptsize 1.0$\uparrow$} & {\scriptsize 0.0018$\uparrow$} & {\scriptsize 1.9$\uparrow$} & {\scriptsize 0.0021$\uparrow$} & {\scriptsize 3.2$\uparrow$} & {\scriptsize 0.0031$\uparrow$} \\
\bottomrule
\end{tabular}
}
\end{table}

\subsection{CIFAR-100}
\label{app:cifar}

Tables~\ref{tab:cifar-r18}--\ref{tab:cifar-wrn} report all 20 detectors on CIFAR-100-C (a 100-class label space disjoint from ImageNet) for the two CIFAR-native backbones, by corruption group and overall Average. e-AURC is reported $\times10^3$. Per column the best detector is in \textbf{bold} and the second/third \underline{underlined}. Each table's bottom row ($\Delta$ vs.\ MSP) is CANARY${-}$MSP ($\uparrow$ = CANARY improves over MSP, $\downarrow$ = worse).

\begin{table}[htbp]
\centering
\caption{\textbf{CIFAR-100 --- ResNet-18.}}
\label{tab:cifar-r18}
\resizebox{\textwidth}{!}{
\begin{tabular}{l cc cc cc cc cc}
\toprule
\multirow{2}{*}{\textbf{Method}} & \multicolumn{2}{c}{\textbf{Noise}} & \multicolumn{2}{c}{\textbf{Blur}} & \multicolumn{2}{c}{\textbf{Weather}} & \multicolumn{2}{c}{\textbf{Digital}} & \multicolumn{2}{c}{\textbf{Average}} \\
\cmidrule(lr){2-3} \cmidrule(lr){4-5} \cmidrule(lr){6-7} \cmidrule(lr){8-9} \cmidrule(lr){10-11}
 & \textbf{e-AURC}$\downarrow$ & \textbf{AUROC}$\uparrow$ & \textbf{e-AURC}$\downarrow$ & \textbf{AUROC}$\uparrow$ & \textbf{e-AURC}$\downarrow$ & \textbf{AUROC}$\uparrow$ & \textbf{e-AURC}$\downarrow$ & \textbf{AUROC}$\uparrow$ & \textbf{e-AURC}$\downarrow$ & \textbf{AUROC}$\uparrow$ \\
\midrule
\multicolumn{11}{l}{\textit{Logit-based}} \\
\midrule
MSP & 161.3 & 0.7095 & 123.4 & \underline{0.7857} & \underline{78.7} & \underline{0.8317} & \underline{90.1} & \underline{0.8184} & 110.2 & \underline{0.7915} \\
MaxLogit & 180.5 & 0.6878 & 136.5 & 0.7637 & 94.5 & 0.8034 & 106.1 & 0.7922 & 126.0 & 0.7667 \\
Energy & 184.9 & 0.6769 & 141.4 & 0.7521 & 99.2 & 0.7918 & 111.2 & 0.7803 & 130.8 & 0.7552 \\
Doctor & 161.6 & 0.7085 & 123.6 & 0.7849 & 79.0 & 0.8305 & 90.3 & 0.8175 & 110.5 & 0.7905 \\
GEN & 183.3 & 0.6790 & 139.9 & 0.7545 & 97.7 & 0.7943 & 109.6 & 0.7828 & 129.2 & 0.7576 \\
\midrule
\multicolumn{11}{l}{\textit{Activation-shaping}} \\
\midrule
ASH & 262.3 & 0.5693 & 261.0 & 0.5994 & 228.8 & 0.6178 & 242.1 & 0.6165 & 247.6 & 0.6028 \\
ReAct & 185.8 & 0.6692 & 150.0 & 0.7336 & 110.5 & 0.7686 & 125.0 & 0.7569 & 140.0 & 0.7363 \\
DICE & 184.1 & 0.6786 & 140.7 & 0.7531 & 98.6 & 0.7928 & 110.7 & 0.7810 & 130.2 & 0.7562 \\
\midrule
\multicolumn{11}{l}{\textit{Feature-based}} \\
\midrule
ViM & 333.2 & 0.4439 & 411.5 & 0.3527 & 410.9 & 0.3141 & 418.4 & 0.3324 & 397.5 & 0.3552 \\
Deep kNN & 340.1 & 0.3989 & 427.4 & 0.3042 & 438.1 & 0.2489 & 443.8 & 0.2700 & 417.2 & 0.2993 \\
MDS & 219.3 & 0.6019 & 206.2 & 0.6248 & 182.8 & 0.6236 & 186.6 & 0.6380 & 197.3 & 0.6234 \\
MDS++ & 165.9 & 0.6736 & 137.4 & 0.7476 & 102.8 & 0.7777 & 111.5 & 0.7738 & 127.0 & 0.7478 \\
RMDS & 170.6 & 0.6556 & 128.2 & 0.7657 & 90.3 & 0.8016 & 97.7 & 0.7983 & 118.4 & 0.7619 \\
RMDS++ & 163.6 & 0.6711 & \underline{122.2} & 0.7775 & 85.4 & 0.8121 & 94.5 & 0.8053 & 113.3 & 0.7729 \\
NNGuide & 173.7 & 0.6935 & 129.7 & 0.7722 & 91.4 & 0.8055 & 104.6 & 0.7919 & 121.6 & 0.7706 \\
fDBD & \textbf{139.6} & \textbf{0.7406} & \textbf{118.0} & \textbf{0.7933} & \underline{77.8} & \underline{0.8325} & \underline{88.4} & \underline{0.8213} & \textbf{103.7} & \textbf{0.8007} \\
TrustScore & \underline{148.7} & 0.7150 & 122.7 & 0.7762 & 82.0 & 0.8200 & 94.3 & 0.8044 & 109.5 & 0.7832 \\
\midrule
\multicolumn{11}{l}{\textit{Softmax-combination}} \\
\midrule
SIRC (L1) & 163.0 & 0.7045 & 125.1 & 0.7817 & 79.5 & 0.8298 & 91.2 & 0.8157 & 111.5 & 0.7881 \\
SIRC (Residual) & 155.4 & \underline{0.7158} & 123.7 & 0.7838 & 79.0 & 0.8304 & 90.7 & 0.8163 & \underline{109.3} & 0.7913 \\
\midrule
\rowcolor{gray!15} \textbf{CANARY (ours)} & \underline{150.3} & \underline{0.7202} & \underline{120.6} & \underline{0.7891} & \textbf{77.0} & \textbf{0.8340} & \textbf{87.9} & \textbf{0.8214} & \underline{106.2} & \underline{0.7959} \\
{\scriptsize ($\Delta$ vs.\ MSP)} & {\scriptsize 11.0$\uparrow$} & {\scriptsize 0.0107$\uparrow$} & {\scriptsize 2.8$\uparrow$} & {\scriptsize 0.0034$\uparrow$} & {\scriptsize 1.7$\uparrow$} & {\scriptsize 0.0023$\uparrow$} & {\scriptsize 2.2$\uparrow$} & {\scriptsize 0.0030$\uparrow$} & {\scriptsize 4.0$\uparrow$} & {\scriptsize 0.0045$\uparrow$} \\
\bottomrule
\end{tabular}
}
\end{table}

\begin{table}[htbp]
\centering
\caption{\textbf{CIFAR-100 --- WideResNet-28-10.}}
\label{tab:cifar-wrn}
\resizebox{\textwidth}{!}{
\begin{tabular}{l cc cc cc cc cc}
\toprule
\multirow{2}{*}{\textbf{Method}} & \multicolumn{2}{c}{\textbf{Noise}} & \multicolumn{2}{c}{\textbf{Blur}} & \multicolumn{2}{c}{\textbf{Weather}} & \multicolumn{2}{c}{\textbf{Digital}} & \multicolumn{2}{c}{\textbf{Average}} \\
\cmidrule(lr){2-3} \cmidrule(lr){4-5} \cmidrule(lr){6-7} \cmidrule(lr){8-9} \cmidrule(lr){10-11}
 & \textbf{e-AURC}$\downarrow$ & \textbf{AUROC}$\uparrow$ & \textbf{e-AURC}$\downarrow$ & \textbf{AUROC}$\uparrow$ & \textbf{e-AURC}$\downarrow$ & \textbf{AUROC}$\uparrow$ & \textbf{e-AURC}$\downarrow$ & \textbf{AUROC}$\uparrow$ & \textbf{e-AURC}$\downarrow$ & \textbf{AUROC}$\uparrow$ \\
\midrule
\multicolumn{11}{l}{\textit{Logit-based}} \\
\midrule
MSP & 168.6 & 0.6746 & 120.6 & 0.7808 & 68.2 & 0.8413 & 83.7 & 0.8237 & 106.4 & 0.7871 \\
MaxLogit & 181.5 & 0.6577 & 126.1 & 0.7695 & 76.6 & 0.8230 & 93.0 & 0.8072 & 115.1 & 0.7714 \\
Energy & 184.6 & 0.6498 & 128.6 & 0.7622 & 79.4 & 0.8148 & 96.3 & 0.7985 & 118.1 & 0.7635 \\
Doctor & 168.9 & 0.6736 & 120.6 & 0.7804 & 68.4 & 0.8403 & 84.0 & 0.8228 & 106.6 & 0.7863 \\
GEN & 183.4 & 0.6514 & 128.0 & 0.7635 & 78.6 & 0.8164 & 95.4 & 0.8001 & 117.2 & 0.7649 \\
\midrule
\multicolumn{11}{l}{\textit{Activation-shaping}} \\
\midrule
ASH & 197.2 & 0.6407 & 134.8 & 0.7590 & 80.1 & 0.8182 & 100.2 & 0.7987 & 123.5 & 0.7617 \\
ReAct & 235.9 & 0.5708 & 208.4 & 0.6136 & 180.6 & 0.6170 & 194.6 & 0.6177 & 202.8 & 0.6071 \\
DICE & 184.5 & 0.6499 & 128.5 & 0.7624 & 79.3 & 0.8149 & 96.2 & 0.7986 & 118.0 & 0.7635 \\
\midrule
\multicolumn{11}{l}{\textit{Feature-based}} \\
\midrule
ViM & 252.8 & 0.5912 & 317.4 & 0.4693 & 293.4 & 0.4524 & 331.8 & 0.4362 & 301.9 & 0.4803 \\
Deep kNN & 337.1 & 0.3902 & 426.1 & 0.2727 & 435.2 & 0.2015 & 450.3 & 0.2212 & 417.2 & 0.2635 \\
MDS & \textbf{139.4} & \underline{0.7044} & \underline{115.5} & 0.7845 & 67.1 & 0.8380 & 80.6 & 0.8236 & \underline{98.1} & 0.7932 \\
MDS++ & \underline{144.9} & 0.6976 & \textbf{114.0} & \underline{0.7868} & \underline{65.4} & 0.8429 & \textbf{79.3} & \underline{0.8275} & \textbf{98.0} & \underline{0.7948} \\
RMDS & 158.7 & 0.6685 & 116.6 & 0.7812 & 67.3 & 0.8378 & 81.7 & 0.8221 & 102.6 & 0.7847 \\
RMDS++ & 148.5 & 0.6899 & \underline{114.9} & \underline{0.7856} & \textbf{65.4} & \underline{0.8430} & \underline{79.3} & \underline{0.8274} & \underline{98.9} & 0.7929 \\
NNGuide & 173.5 & 0.6696 & 126.0 & 0.7710 & 76.1 & 0.8246 & 91.9 & 0.8086 & 113.1 & 0.7750 \\
fDBD & \underline{145.2} & \textbf{0.7094} & 116.9 & \textbf{0.7898} & \underline{66.0} & \textbf{0.8462} & \underline{79.5} & \textbf{0.8302} & 99.0 & \textbf{0.7996} \\
TrustScore & 148.7 & 0.6870 & 118.0 & 0.7791 & 66.5 & 0.8391 & 80.4 & 0.8233 & 100.4 & 0.7885 \\
\midrule
\multicolumn{11}{l}{\textit{Softmax-combination}} \\
\midrule
SIRC (L1) & 170.2 & 0.6700 & 121.7 & 0.7779 & 68.8 & 0.8396 & 84.8 & 0.8211 & 107.4 & 0.7843 \\
SIRC (Residual) & 154.2 & \underline{0.7031} & 119.6 & 0.7814 & 67.2 & \underline{0.8431} & 82.9 & 0.8251 & 102.8 & \underline{0.7938} \\
\midrule
\rowcolor{gray!15} \textbf{CANARY (ours)} & 164.8 & 0.6781 & 119.7 & 0.7815 & 67.6 & 0.8418 & 83.4 & 0.8239 & 105.2 & 0.7882 \\
{\scriptsize ($\Delta$ vs.\ MSP)} & {\scriptsize 3.8$\uparrow$} & {\scriptsize 0.0035$\uparrow$} & {\scriptsize 0.8$\uparrow$} & {\scriptsize 0.0007$\uparrow$} & {\scriptsize 0.6$\uparrow$} & {\scriptsize 0.0006$\uparrow$} & {\scriptsize 0.3$\uparrow$} & {\scriptsize 0.0002$\uparrow$} & {\scriptsize 1.2$\uparrow$} & {\scriptsize 0.0011$\uparrow$} \\
\bottomrule
\end{tabular}
}
\end{table}

\clearpage

\section{SIRC per-backbone behavior}
\label{app:sirc-backbone}

SIRC (Residual)'s strength is concentrated rather than broad. Almost all of its top-1 cells---31 of 35 (e-AURC) and 51 of 56 (AUROC)---fall on ResNet-50 alone, where it and CANARY tie within seed noise on both metrics. On the other four backbones it tops at most four cells.

\begin{table}[htbp]
\centering
\caption{\textbf{SIRC per-backbone behavior.} ``Top-1 Cells'' = per-backbone single-best count of the 20 detectors (e-AURC / AUROC). Showing both metrics exposes the seed-noise-level gaps---on ResNet-50, SIRC (Residual) and CANARY tie on both, and even disagree on which metric each tops.}
\label{tab:sirc-backbone}
\resizebox{0.9\textwidth}{!}{
\begin{tabular}{l ccc ccc ccc}
\toprule
\multirow{2}{*}{\textbf{Backbone}} & \multicolumn{3}{c}{\textbf{MSP}} & \multicolumn{3}{c}{\textbf{CANARY}} & \multicolumn{3}{c}{\textbf{SIRC (Residual)}} \\
\cmidrule(lr){2-4} \cmidrule(lr){5-7} \cmidrule(lr){8-10}
& \textbf{e-AURC} $\downarrow$ & \textbf{AUROC} $\uparrow$ & \textbf{Top-1 Cells} & \textbf{e-AURC} $\downarrow$ & \textbf{AUROC} $\uparrow$ & \textbf{Top-1 Cells} & \textbf{e-AURC} $\downarrow$ & \textbf{AUROC} $\uparrow$ & \textbf{Top-1 Cells} \\
\midrule
ViT-B/16        & 58.2 & 0.8537 & 3 / 3 & 57.8 & 0.8545 & 67 / 68 & 60.1 & 0.8471 & 4 / 2 \\
ResNet-50       & 82.3 & 0.8389 & 0 / 0 & 81.3 & 0.8401 & 33 / 15 & 81.2 & 0.8407 & 31 / 51 \\
Swin-B          & 74.2 & 0.8298 & 0 / 0 & 66.6 & 0.8377 & 61 / 59 & 69.7 & 0.8316 & 0 / 2 \\
ConvNeXt-B      & 61.4 & 0.8531 & 0 / 0 & 56.9 & 0.8576 & 74 / 73 & 59.3 & 0.8533 & 0 / 1 \\
EfficientNet-B4 & 87.5 & 0.8318 & 0 / 0 & 78.0 & 0.8406 & 73 / 74 & 83.1 & 0.8317 & 0 / 0 \\
\bottomrule
\end{tabular}
}
\end{table}

That concentration is the point: a global residual norm helps only where an architecture's failure modes happen to align with it, whereas CANARY's per-class residual isolates a signal that transfers---top-1 cells on every backbone and never regressing past the seed-noise floor.

\section{The Headroom Law on ObjectNet}
\label{app:headroom-objectnet}

Figure~\ref{fig:headroom-law} is shown in the body on the ImageNet-C/-A/-R panels. Figure~\ref{fig:headroom-objectnet} confirms it reproduces on ObjectNet on the same forecastable validation MSP $x$-axis.

\begin{figure}[h]
\centering
\includegraphics[width=0.50\textwidth]{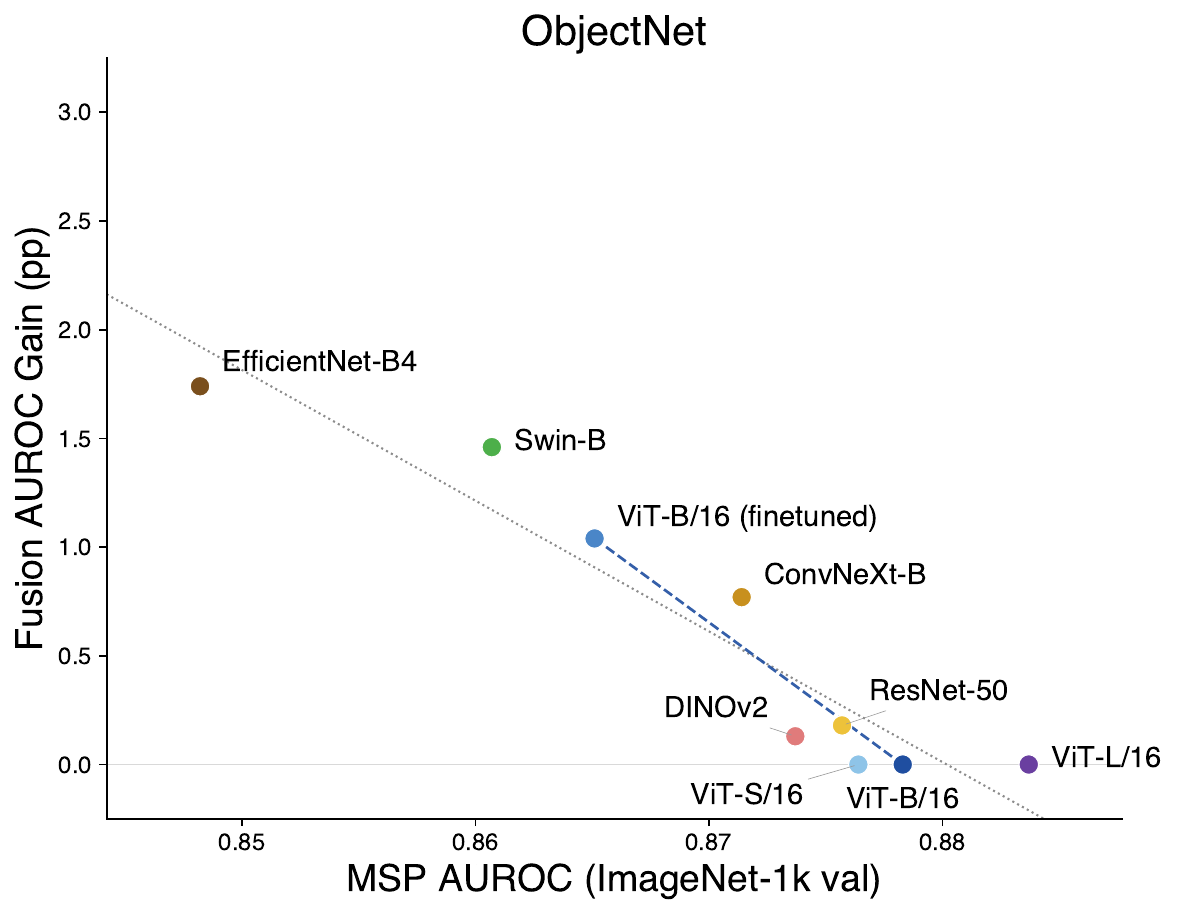}
\caption{\textbf{The headroom law on ObjectNet.} CANARY's fusion AUROC gain over MSP on ObjectNet, plotted against each backbone's ImageNet-1k validation MSP AUROC. The same negative slope as the Figure~\ref{fig:headroom-law} panels reproduces here.}
\label{fig:headroom-objectnet}
\end{figure}

\newpage
\section{Comparison with Kernel PCA}
\label{app:kpca}
Kernel PCA (KPCA)~\cite{NEURIPS2024_f2543511} detects out-of-distribution inputs by measuring the reconstruction error of features projected onto a non-linear principal subspace. To compare it with CANARY, we evaluate KPCA in three distinct roles: as a standalone failure detector, fused with MSP, and as an alternative basis for CANARY's residual. These evaluations span four diverse backbones: ViT-B/16, ResNet-50, Swin-B, and the self-supervised DINOv2. Following the original method, we implement its two explicit-feature-map kernels---cosine (CoP; $\phi(z)=z/\lVert z\rVert$) and cosine-Gaussian (CoRP; using random Fourier features with $M{=}4096$). We fit PCA on the correctly classified in-distribution validation features, retaining components that explain $90\%$ of the variance, and score each input by its reconstruction error. All reported results represent the mean AUROC across 77 shift cells, under the same out-of-sample protocol used in our main experiments.

\vspace{0.7em}
\noindent \textbf{KPCA as a failure detector.} On its own, KPCA reconstruction error provides a weak failure signal ($0.68$--$0.75$ AUROC), falling well below the MSP baseline (Table~\ref{tab:kpca-detector}). When fused with MSP using CANARY's percentile calibration and weighted-Fisher formulation, KPCA regresses below MSP in 6 of 8 cases (by up to $0.027$ AUROC) and only marginally exceeds it in two (CoP on Swin-B and DINOv2, by at most $0.0035$). Furthermore, once fused, the non-linear CoRP kernel consistently trails CoP across all backbones. In contrast, CANARY improves upon MSP across all four backbones, with DINOv2 being the sole instance where CoP$+$MSP attains a slightly higher mean than CANARY ($0.8553$ vs.\ $0.8538$). Since both methods share an identical fusion rule, this discrepancy isolates the quality of the underlying signal: a class-agnostic, whole-feature reconstruction error is largely redundant with MSP (Section~\ref{sec:analysis-redundancy}). Consequently, its contribution under fusion is small and inconsistent in sign across backbones.

\begin{table}[h]
\centering
\caption{\textbf{KPCA reconstruction error as a failure detector.} Evaluated standalone and fused with MSP via CANARY's percentile calibration and weighted-Fisher formulation. Mean AUROC over 77 shift cells; best per row in \textbf{bold}.}
\label{tab:kpca-detector}
\resizebox{0.9\linewidth}{!}{
\begin{tabular}{lcccccc}
\toprule
\textbf{Backbone} & \textbf{MSP} & \textbf{CoP} & \textbf{CoRP} & \textbf{CoP $+$ MSP} & \textbf{CoRP $+$ MSP} & \textbf{CANARY (ours)} \\
\midrule
ViT-B/16  & 0.8537 & 0.6945 & 0.7158 & 0.8492 & 0.8305 & \textbf{0.8545} \\
ResNet-50 & 0.8389 & 0.6794 & 0.6833 & 0.8385 & 0.8124 & \textbf{0.8401} \\
Swin-B    & 0.8298 & 0.7265 & 0.7545 & 0.8333 & 0.8275 & \textbf{0.8377} \\
DINOv2    & 0.8519 & 0.7086 & 0.7397 & \textbf{0.8553} & 0.8486 & 0.8538 \\
\bottomrule
\end{tabular}
}
\end{table}

\vspace{0.7em}
\noindent \textbf{KPCA's native fusion rule.} Following~\cite{guan2023revisit}, KPCA~\cite{NEURIPS2024_f2543511} multiplicatively combines the reconstruction error with a logit score via $(1-\tilde{e})\cdot S_{\mathrm{logit}}$, where $\tilde{e}$ is the in-distribution percentile of the error. When applied to MSP, this rule substantially degrades performance, dropping AUROC by $5.6$--$10.6$ pp on every backbone (Table~\ref{tab:kpca-fusion}); its Energy-based variant similarly falls $6.8$--$14.8$ pp below MSP. We attribute this to the rule's design: built to separate in-distribution from OOD inputs, the multiplicative factor rescales MSP by a class-agnostic error, thereby perturbing the within-distribution confidence ranking on which failure prediction relies.

\begin{table}[h]
\centering
\caption{\textbf{KPCA's multiplicative fusion.} The rule of~\cite{guan2023revisit,NEURIPS2024_f2543511} applied to MSP, compared against CANARY. Mean AUROC across 77 shift cells; best per column in \textbf{bold}. $\tilde{e}$ denotes the in-distribution percentile of the reconstruction error.}
\label{tab:kpca-fusion}
\resizebox{0.7\linewidth}{!}{
\begin{tabular}{lcccc}
\toprule
\textbf{Method} & \textbf{ViT-B/16} & \textbf{ResNet-50} & \textbf{Swin-B} & \textbf{DINOv2} \\
\midrule
MSP                              & 0.8537 & 0.8389 & 0.8298 & 0.8519 \\
$(1-\tilde{e})\cdot$MSP, CoP     & 0.7521 & 0.7580 & 0.7661 & 0.7734 \\
$(1-\tilde{e})\cdot$MSP, CoRP    & 0.7480 & 0.7405 & 0.7736 & 0.7758 \\
\midrule
\textbf{CANARY (ours)}           & \textbf{0.8545} & \textbf{0.8401} & \textbf{0.8377} & \textbf{0.8538} \\
\bottomrule
\end{tabular}
}
\end{table}

\vspace{0.7em}
\noindent \textbf{KPCA as CANARY's basis.} To evaluate KPCA as a representation basis, we replace CANARY's linear PCA basis with the cosine-kernel basis (CoP's explicit feature map, \ie, PCA on $\ell_2$-normalized features). Holding the per-class kNN, calibration, and fusion constant, this substitution changes CANARY's AUROC by at most $0.0007$ (Table~\ref{tab:kpca-basis}). The kernel's non-linearity thus yields no tangible benefit, consistent with kNN's reliance on local neighborhood geometry rather than on a particular choice of coordinates. This result extends the basis invariance of Section~\ref{sec:analysis-basis} (PCA, AE, and VAE) to a cosine-kernel basis.

\begin{table}[h]
\centering
\caption{\textbf{CANARY with a cosine-kernel (CoP) basis.} Mean AUROC over 77 shift cells.}
\label{tab:kpca-basis}
\resizebox{0.6\linewidth}{!}{
\begin{tabular}{lccc}
\toprule
\textbf{Backbone} & \textbf{Linear-PCA basis} & \textbf{CoP basis} & \textbf{$\Delta$} \\
\midrule
ViT-B/16  & 0.8545 & 0.8543 & $-0.0002$ \\
ResNet-50 & 0.8401 & 0.8399 & $-0.0002$ \\
Swin-B    & 0.8377 & 0.8370 & $-0.0007$ \\
DINOv2    & 0.8538 & 0.8541 & $+0.0003$ \\
\bottomrule
\end{tabular}
}
\end{table}

\newpage
\section{Model scale: ViT-S/16 and ViT-L/16}
\label{app:vit-scale}

ViT-S/16 and ViT-L/16 probe the model-scale axis of Section~\ref{sec:result-generality}. As the smaller and larger siblings of the primary ViT-B/16, they differ only in capacity. They sit outside the main tables, which span distinct architectures, and are therefore reported here. Capacity monotonically raises the clean-validation MSP quality, and CANARY's behavior tracks the resulting headroom.

\begin{table}[htbp]
\centering
\caption{\textbf{Performance across the ViT model scales.} Shift AUROCs are the mean across all shift cells; Val MSP is the clean-validation AUROC. $\dagger$ indicates that CANARY outperforms MSP on every shift cell.}
\label{tab:vit_scale}
\resizebox{0.8\linewidth}{!}{
  \begin{tabular}{l c cc ccc}
  \toprule
  \multirow{2}{*}{\textbf{Backbone}} & \textbf{Val MSP} & \multicolumn{2}{c}{\textbf{Shift
  AUROC$\uparrow$}} & \multicolumn{3}{c}{\textbf{CANARY vs.\ MSP}} \\
  \cmidrule(lr){3-4} \cmidrule(lr){5-7}
  & \textbf{AUROC$\uparrow$} & \textbf{MSP} & \textbf{CANARY} & \textbf{$\Delta$ (pp)} &
  \textbf{$\ge$ MSP Cells} & \textbf{Max Drop} \\
  \midrule
  ViT-S/16        & 0.8764 & 0.8458 & 0.8469 & $+0.11$ & 77/77 & $+0.0002^\dagger$ \\
  ViT-B/16 (main) & 0.8783 & 0.8537 & 0.8545 & $+0.08$ & 74/77 & $-0.0002$ \\
  ViT-L/16        & 0.8837 & 0.8563 & 0.8573 & $+0.10$ & 66/77 & $-0.0005$ \\
  \bottomrule
  \end{tabular}
}
\end{table}
  
On the smaller ViT-S, the residual carries a complementary failure signal, allowing CANARY to improve over MSP on all 77 shift cells, thereby maintaining strict baseline dominance at a lower capacity. On ViT-L, capacity pushes MSP to the accuracy ceiling; the per-class kNN residual is already correlated with MSP on validation (Pearson $+0.52$), and its standalone AUROC (0.7523, comparable to ViT-B's 0.7465) confirms that the pipeline is sound. Rather, the residual is less consistent in providing novel information beyond what MSP already captures. Consequently, while ViT-L's mean improvement ($+0.10$ pp) is comparable to that of ViT-S, the gains are no longer uniform; CANARY marginally trails MSP on 11 of 77 cells, but the worst-case drop ($-0.0005$ AUROC) falls within seed noise, rendering it an effective tie.

\fi
\end{document}